\documentclass[10pt,twocolumn,letterpaper]{article}

\usepackage[pagenumbers]{cvpr} %

\usepackage{multirow}
\usepackage{booktabs}
\usepackage{amssymb} %
\usepackage{overpic}

\usepackage[accsupp]{axessibility}

\newcommand{\parag}[1]{{\noindent\textbf{#1}}}

\newcommand{\shortname}{\emph{PoseTraj}\xspace}
\newcommand{\shortdname}{\emph{PoseTraj-10K}\xspace}

\definecolor{cvprblue}{rgb}{0.21,0.49,0.74}
\usepackage[pagebackref,breaklinks,colorlinks,citecolor=cvprblue]{hyperref}

\title{\shortname: Pose-Aware Trajectory Control in Video Diffusion}

\author{
    Longbin Ji$^1$ \quad Lei Zhong$^1$ \quad Pengfei Wei$^2$ \quad Changjian Li$^1$ \vspace{2mm}\\
    $^1$University of Edinburgh \quad $^2$Nanyang Technological University \vspace{1mm}\\
    \href{https://robingg1.github.io/Pose-Traj/}{https://robingg1.github.io/Pose-Traj/}
}

\begin{document}

\twocolumn[{
\renewcommand\twocolumn[1][]{#1}
\maketitle 
\begin{center}
\vspace{-6mm}
    \centering
    \includegraphics[width=1\textwidth]{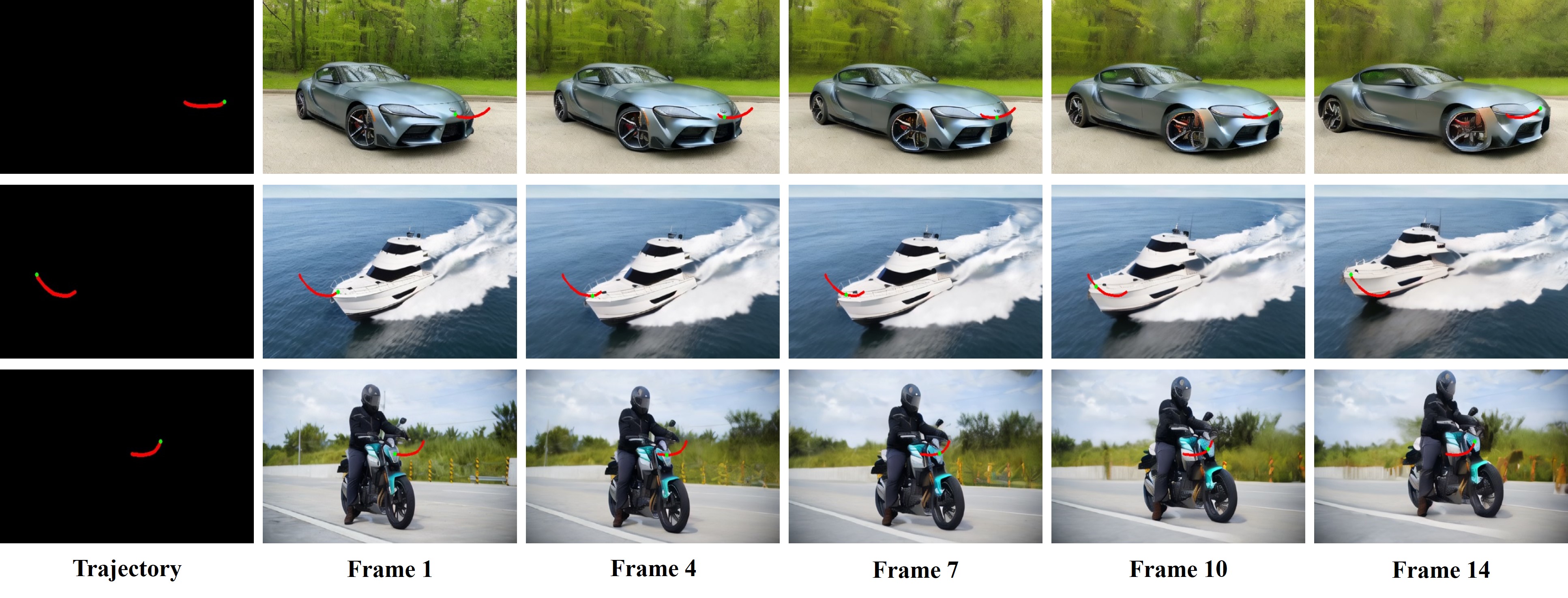}
    \captionsetup{type=figure}
    \vspace{-7mm}
    \captionof{figure}{
    \shortname produces plausible dragging videos, where objects follow a rotational trajectory with awareness of changing poses.
    }
    \label{fig:teaser}
\end{center}
}]

\begin{abstract}
Recent advancements in trajectory-guided video generation have achieved notable progress. 
However, existing models still face challenges in generating object motions with potentially changing 6D poses under wide-range rotations, due to limited 3D understanding. 
To address this problem, we introduce \shortname, a pose-aware video dragging model for generating 3D-aligned motion from 2D trajectories. 
Our method adopts a novel two-stage pose-aware pretraining framework, improving 3D understanding across diverse trajectories. 
Specifically, we propose a large-scale synthetic dataset \shortdname, containing 10k videos of objects following rotational trajectories, and enhance the model perception of object pose changes by incorporating 3D bounding boxes as intermediate supervision signals. 
Following this, we fine-tune the trajectory-controlling module on real-world videos, applying an additional camera-disentanglement module to further refine motion accuracy.
Experiments on various benchmark datasets demonstrate that our method not only excels in 3D pose-aligned dragging for rotational trajectories but also outperforms existing baselines in trajectory accuracy and video quality.

\end{abstract}
    
\section{Introduction}

We address the problem of controlling video generation with trajectories involving potential object rotations, as illustrated in Fig.~\ref{fig:teaser}.
Recently, significant advancements have been made in text-to-video generation~\cite{esser2023structure,chen2023videocrafter1,he2022lvdm}. Building on these foundational models, many works have explored the use of additional guidance signals, such as images~\cite{blattmann2023stable, ni2023conditional, ni2024ti2v}, trajectories~\cite{yin2023dragnuwa, wu2024draganything, zhang2024tora}, audio \cite{tian2024emo, xu2024hallo}, and poses~\cite{ma2024follow, hu2024animate, wang2024unianimate, zhu2024champ} to control the content in the generated video.
Among these, trajectory-guided motion control~\cite{yin2023dragnuwa, wu2024draganything} has been gaining increasing attention due to its interactive and user-friendly nature.

\begin{figure*}[!ht]
\centering
\includegraphics[width=0.95\textwidth]{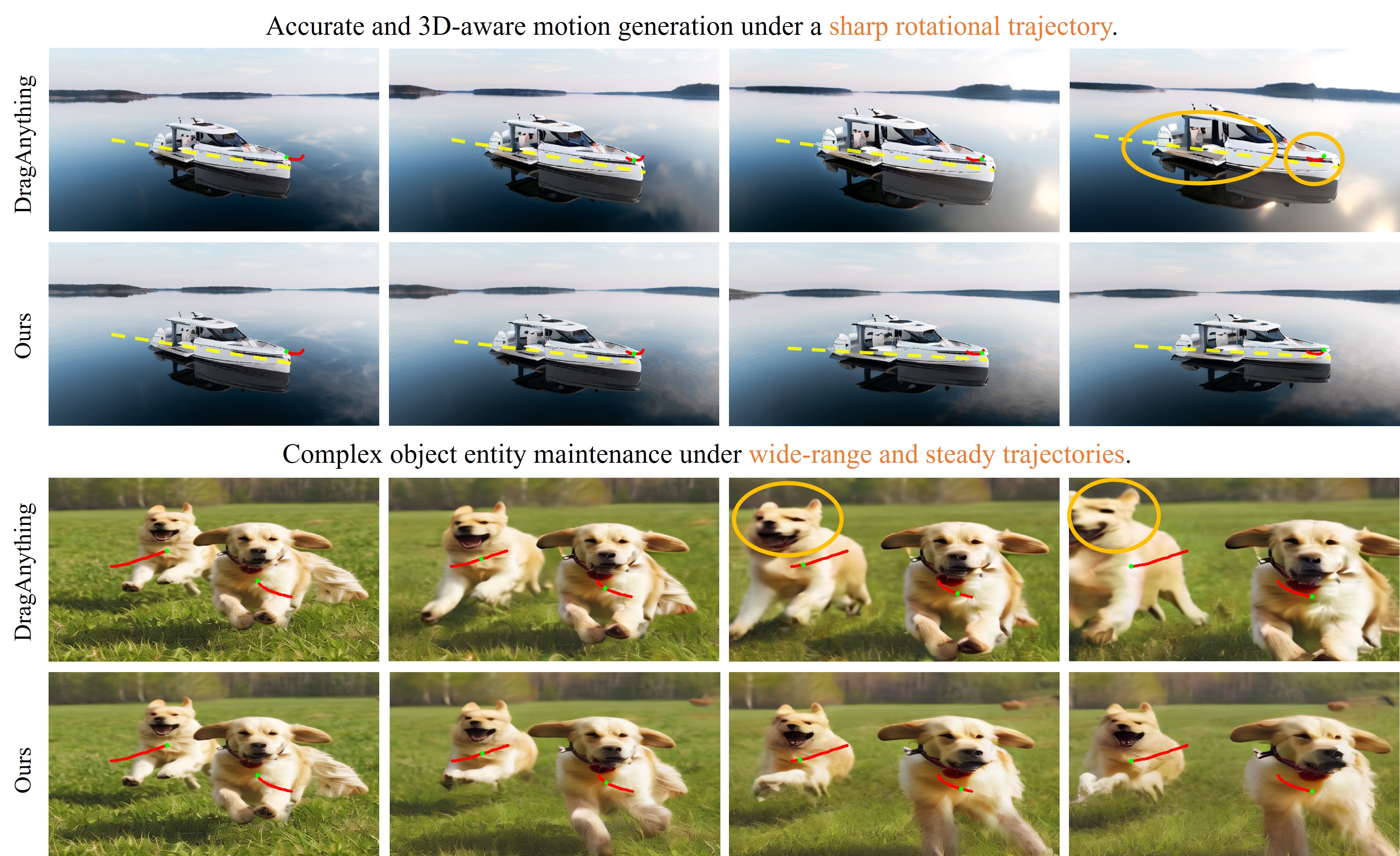}
\vspace{-2mm}
\caption{The video generation performance of DragAnything $vs.$ Ours under complex trajectories on static and dynamic objects. The yellow dashed line indicates the boat's orientation and helps perceive object rotation, while the orange circle highlights visual defects.}
\label{fig:intro_demo} 
\vspace{-4mm}
\end{figure*}

Trajectory-guided motion control requires the target object to follow a specified path accurately, preserving temporal coherence and realism in the generated video.
Earlier works IPoke~\cite{blattmann2021ipoke} and MCDiff~\cite{chen2023motion} leverage dragging animation only in the limited human domain. 
DragNUWA \cite{yin2023dragnuwa} is the first to attempt trajectory control in open-domain videos. 
DragAnything~\cite{wu2024draganything} further enhances object-level motion awareness by extracting object entities using corresponding masks and integrating them with trajectory guidance. %
However, these works constrain the object to follow the trajectory in a 2D image space, without considering potential changes of the object's 6D pose (position and orientation). 
Consequently, they can perform well in translational motion but fail to control objects when trajectories implicitly involve rotations (see Fig. \ref{fig:intro_demo}).
These limitations arise from two key factors: i) in the collected video datasets, object motions are more commonly translational, while rotations are rare and difficult to annotate automatically, ii) it is inherently ill-posed to infer potential rotations from 2D trajectories and objects in pixel space.

In this paper, we present a novel pose-aware, trajectory-guided motion control framework, dubbed \shortname, which can perceive potential rotations in the trajectory and perform corresponding control (as shown in Fig.~\ref{fig:teaser}).
The ability to perceive rotations in the trajectory originates from a two-stage pose-aware pre-training pipeline. 
Specifically, we propose a specialized synthetic dataset \shortdname, containing over 10,000 videos with diverse trajectories for 2,000 distinct objects involving rotations, paired with the objects' 3D bounding boxes. 
This dataset circumvents the challenge of accurately estimating trajectories and 3D bounding boxes from open-world videos.
With this synthetic dataset, in the first stage, we pre-train the model to generate video frames with accurate 3D bounding boxes following trajectories, without emphasizing appearance details.
The introduction of 3D bounding box generation helps the model understand the object location along the driving trajectories and makes it aware of potential rotational changes. 
In the second stage, 
we further pre-train the model to focus on refining object appearance with the learned knowledge of object entities and poses.
After the two-stage pre-training, the model can perceive potential rotations and maintain stable object entities along the trajectory.
Finally, we perform camera-disentangled fine-tuning on a real-world video dataset to enhance the model's generalization to real-world videos.

Experiments on the VIPSeg benchmark dataset demonstrate that \shortname achieves substantial improvements over DragNUWA and DragAnything in both quantitative metrics and qualitative visual performance. As illustrated in Fig.~\ref{fig:intro_demo}, our model provides more accurate motion tracking and superior object reconstruction quality, especially in scenarios involving complex rotations or dynamic objects.

To summarize, our contributions are:
\begin{itemize}
    \item We propose a novel pose-aware pre-training process that employs 3D bounding boxes as intermediate supervision to enhance the model’s 6D pose understanding. This capability is further generalized to open-domain video through camera-disentangled fine-tuning.
    
    \item We introduce \shortdname, the first synthetic dataset augmented with both trajectories and 3D bounding boxes, featuring diverse animated objects undergoing wide-range rotational motions along complex paths.
    
    \item Experimental results show that \shortname outperforms SoTA baselines in trajectory-following accuracy and video realism when dragging objects along a path, particularly in scenarios where rotations are included.
\end{itemize}

\section{Related Work}
\subsection{Controllable Video Generation Model}

\textbf{Video generation model.} While image generation models~\cite{ramesh2022hierarchical,ramesh2022hierarchical,saharia2022photorealistic, panfusion2024} have achieved impressive results, video generation remains a relatively nascent field.  
Early works such as NUWA~\cite{wu2022nuwa} improved video quality by unifying representations across multimodal learning. CogVideo~\cite{hong2022cogvideo} was the first to fine-tune a text-to-video model using pre-trained text-to-image weights with minimal additional parameters. Make-A-Video~\cite{singer2022make} further advanced efficiency by introducing pseudo-3D convolutional layers and temporal attention mechanisms.  
To mitigate the computational intensity of pixel-space diffusion, recent video diffusion models~\cite{blattmann2023videoldm, he2022lvdm, hu2024animate} have adopted latent diffusion methods~\cite{rombach2022high}. With access to larger volumes of high-quality video data, state-of-the-art models such as VideoCrafter~\cite{chen2023videocrafter1}, Gen-2~\cite{esser2023structure}, SVD~\cite{blattmann2023stable}, and SORA~\cite{brooks2024video} have demonstrated remarkable realism and high-fidelity video generation.

\noindent \textbf{Trajectory-guided video generation.}
Recently, trajectory-guided video generation has gained significant attention due to its user-friendly and interactive nature. Early trajectory-based animation methods, such as IPoke~\cite{blattmann2021ipoke} and MCDiff~\cite{chen2023motion}, were limited to human motion synthesis, relying on skeleton-based conditioning.  
DragNUWA~\cite{yin2023dragnuwa} was the first to introduce dragging control for open-domain videos by fine-tuning the pre-trained image-to-video model SVD with trajectory guidance from large-scale real-world videos. However, maintaining temporal consistency during large motions remains a challenge, often leading to trajectory-following inaccuracies.

To address this challenge, DragAnything~\cite{kirillov2023segment} enhanced object identity awareness by leveraging a pre-trained diffusion-based image encoder for explicit entity extraction from object segmentation masks. 
Tora~\cite{zhang2024tora} further improved trajectory guidance by integrating fusion layers into each DiT basic block within a scalable architecture.
Other works, such as Direct-A-Video~\cite{yang2024direct} and MotionCtrl~\cite{wang2023motionctrl}, aimed to disentangle object motion learning from passive camera movement, enabling more fine-grained control. 
However, existing models lack strong pose-awareness for handling 3D motion along rotational trajectories. Our work addresses this limitation by explicitly incorporating pose-aware pretraining to enhance the model’s understanding of object pose variations.

\subsection{Training on Synthetic Dataset}
Due to the scarcity of comprehensive, large-scale real-world data in certain domains, many studies have leveraged synthetic datasets to enhance model performance for specific tasks.
Focusing on fine-grained 3D reconstruction with diverse materials, OpenMaterial~\cite{dang2024openmaterial} introduced a dataset of 1,001 3D objects, each manually rendered under varying materials and lighting conditions. For image decomposition, the intrinsic diffusion model was trained on the synthetic dataset InteriorVerse~\cite{zhu2022learning}, which provides high-quality intrinsic rendering images. X-RGB~\cite{zeng2024rgb} further developed a unified framework for image decomposition and synthesis, leveraging self-rendered synthetic image pairs conditioned on various intrinsic modalities.  
In controllable video generation, PuppetMaster~\cite{li2024puppet} recently introduced a part-level animation model, rendering high-quality animatable objects sampled from Objaverse~\cite{deitke2024objaverse}.

For our pose-aware pretraining, we propose a synthetic dataset with complex trajectories and precisely annotated 3D bounding boxes, enabling the animation of diverse objects along various wide-range rotational trajectories.

\begin{figure}[!b]
\vspace{-0.4cm}
\centering
\includegraphics[width=0.97\linewidth]{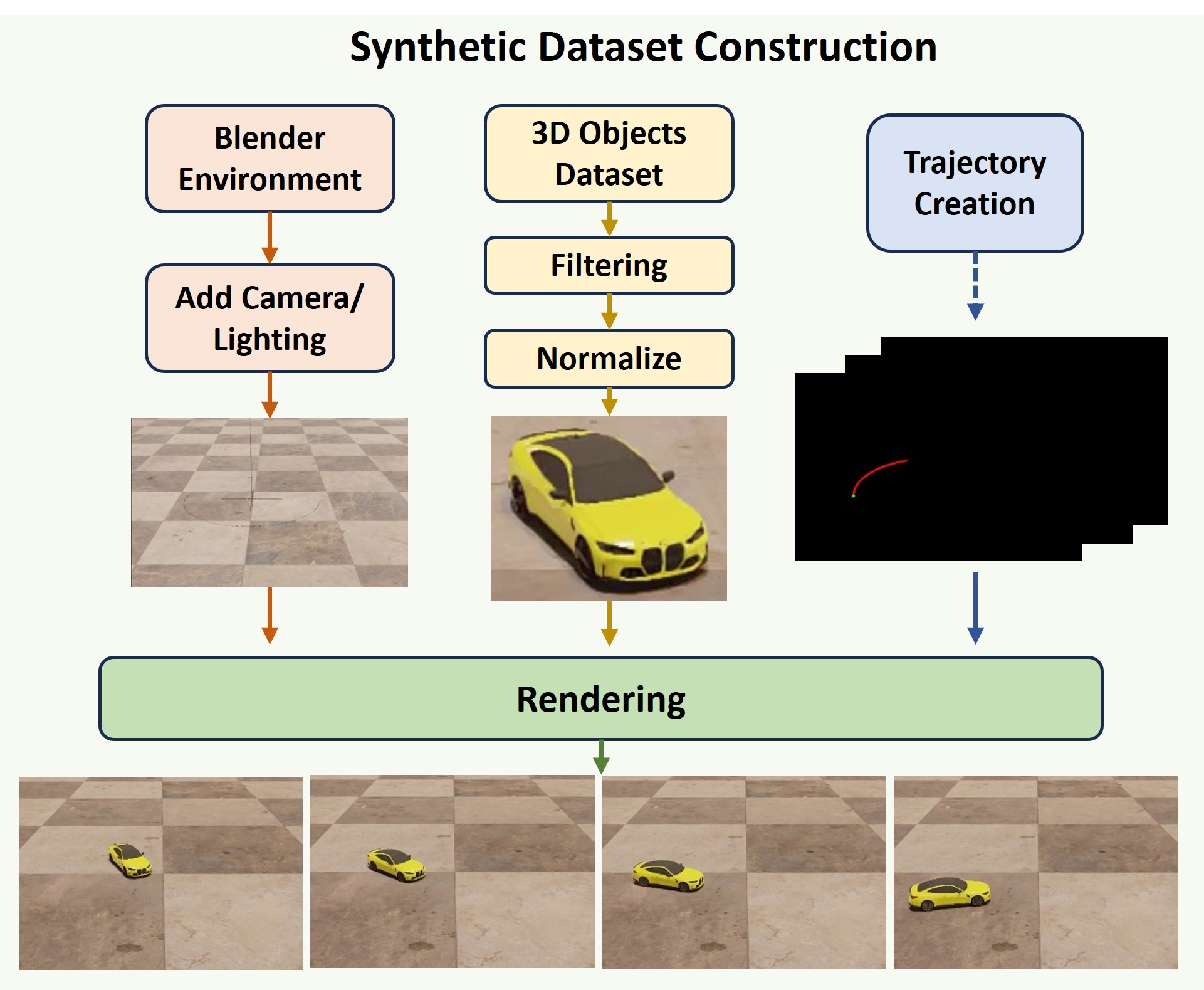}
\vspace{-3mm}
\caption{Data construction pipeline of our synthetic dataset \shortdname. The whole configuration, including environment setup, object sampling, and trajectory sampling, is displayed.}
\label{fig:data_construct} 
\end{figure}

\section{\shortdname Dataset Construction}
\label{sec:dataset}

Training a trajectory-guided video generation model requires paired data, \ie, high-quality video with clearly identifiable moving objects and precisely annotated trajectories. 

However, obtaining such data from open-domain videos presents several significant challenges.
First, complex motions, such as wide-range rotations, are typically rare in filtered video datasets and difficult to annotate accurately using point-tracking or flow-matching estimators. 
Moreover, large motions are often tightly coupled with camera movements, making it challenging to disentangle object motion from camera motion. 
As a result, obtaining precise 6D poses of moving objects from open-domain videos remains infeasible with current pose estimation methods \cite{ornek2025foundpose, caraffa2024freeze}.

\begin{figure*}[!ht]
\centering
    \begin{overpic}[width=0.98\textwidth]{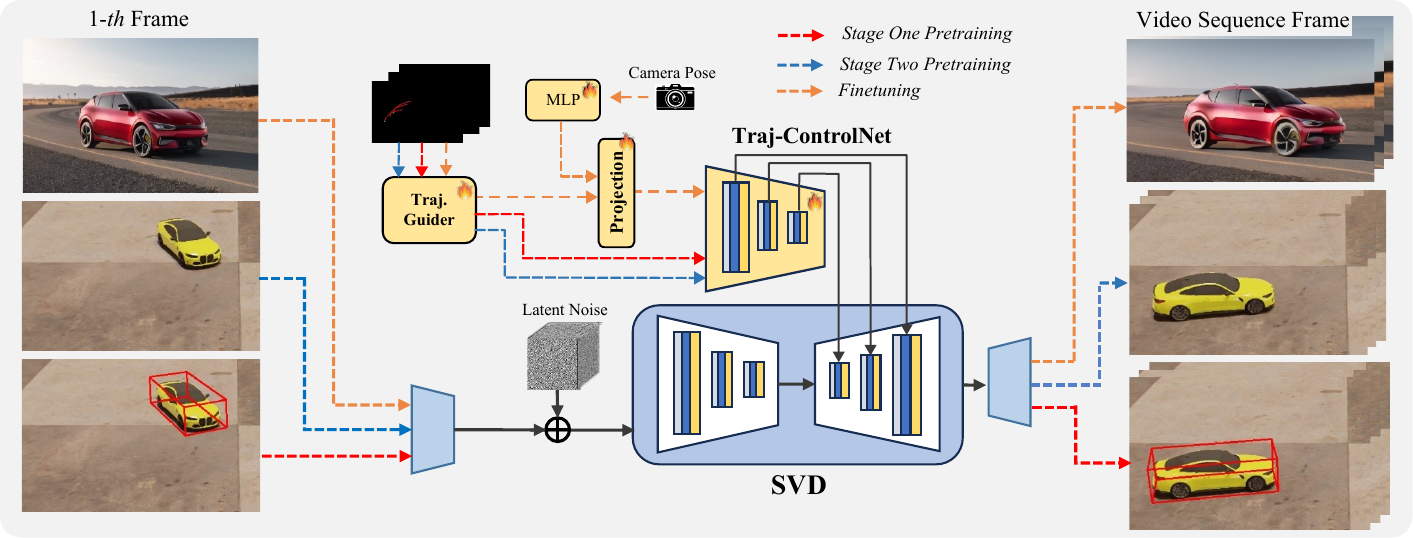}
        \put(30.2,7) {\small $\mathcal{E}$}
        \put(70.7,10.5) {\small $\mathcal{D}$}
    \end{overpic}
    \vspace{-2mm}
    \caption{Method overview. Our \shortname first utilizes two-stage pose-aware pre-training on our synthetic dataset to obtain 3D-enhanced awareness for rotational trajectory-following capacity and further exploits camera-disentangled finetuning to adapt this ability on open-domain videos. The dashed colorful arrows demonstrate the dedicated data flows during the three training stages, while the black arrows are shared in all stages. All yellow blocks are trainable, while other blocks are frozen.
    }
    \label{fig:pipeline}
    \vspace{-4mm}
\end{figure*}

We thus construct a synthetic dataset \shortdname, that offers three key advantages:  
i) it provides a diverse and comprehensive range of objects, enhancing robust motion understanding, particularly for complex geometric structures and wide-range rotations;
ii) it enables the creation of a wide variety of precise trajectories, ensuring motion consistency without interference from camera movements;
iii) through controlled rendering, we can effortlessly obtain precise 3D bounding box parameters, which serve as crucial intermediate supervision signals in our pretraining process.

\cref{fig:data_construct} illustrates our rendering pipeline implemented in Blender.  
First, we set up a realistic virtual scene and sampled 2,000 high-quality 3D models from Objaverse~\cite{deitke2024objaverse}. %
To ensure the quality of the selected models, we employ GPT-4v~\cite{GPT:24} to filter out overly complex or uncommon objects, followed by a manual selection, resulting in the top 2,000 models that best represent high-quality daily objects. 
Each model is normalized and assigned a unique random trajectory with randomized rotational angles, trajectory shape, and length. During the animation, the 3D model follows its designated trajectory while maintaining a rotation center, enabling rotation-focused motion for the final rendering. 
\cref{fig:data_construct} bottom displays four frames of the rendered video, check out our dataset and the supplementary for more examples and further data construction details.

\section{Method}
\label{sec:method}
Given an input image $\mathbf{I} \in \mathbb{R}^{H \times W \times 3}$ containing foreground objects and a driving trajectory 
$\mathbf{tr} = \{(x_i, y_i)\}_{i=1}^L$ drawn as a set of line segment images $\mathbf{I_{tr}} = \{I_{tr}^{i}\}_{i=1}^{L}$ (with last empty padding), our goal is to generate a sequence of animated video frames 
$\{f_i\}_{i=1}^{L}$, where the relevant object accurately follows the trajectory, while maintaining a coherent and natural video appearance.

Fig.~\ref{fig:pipeline} presents an overview of our approach, which is built upon a latent stable video diffusion model, SVD~\cite{blattmann2023stable}.
Our base framework mainly consists of three components: a latent diffusion model \( \epsilon_\theta \) (3D U-Net~\cite{ronneberger2015u}) to denoise the latent noise, and an encoder ($\mathcal{E}$) - decoder ($\mathcal{D}$) pair to compress video frames into the latent space and reconstruct the denoised latent features back into video frames.
Inspired by the success of ControlNet~\cite{ni2023conditional} in guiding the pre-trained diffusion model, we design Traj-ControlNet, which consists of a trainable copy of the encoder blocks of SVD.  
Given a sequence of trajectory images \( \mathbf{I_{tr}} \), we adopt a trajectory guider - a 3D ConvNet, to encode trajectory features, which is then fed into the Traj-ControlNet to predict the residual features for SVD, enabling the generated video to follow the trajectory. 
In the following subsections, we detail on the novel technical designs to achieve our goal.

\subsection{Two-stage Pose-aware Pretraining}
\label{section:pretraining}
To enhance the model’s ability to perceive potential 6D pose changes within the 2D-pixel space, we select the 3D bounding box (bbox)~\cite{wang2024boximator} as the primary 3D supervision signals, as they provide precise positional and pose information. %
We propose a novel two-stage pose-aware pretraining pipeline that incorporates 3D bbox generation as an intermediate supervision signal. The two pre-training is on our \shortdname synthetic dataset.

\noindent\textbf{Stage One: 3D bbox-guided localization.}  
Instead of separately regressing 3D bbox parameters and generating the object's appearance for each video frame, we simultaneously generate 3D bboxes in the image space along with the objects, as shown in Fig.~\ref{fig:pipeline} (red arrows). 
To this end, given one data example in our dataset, we first render the 3D bounding box into the pixel space on top of the image to obtain $\mathbf{I}_{bbox}$ (the bottom-left figure in \cref{fig:pipeline}). We apply the same rendering process for other frames to obtain the bounding box augmented video for supervision. 
This joint generation process helps the model better understand the object's location along the driving trajectory and makes it aware of potential rotational changes.

\noindent\textbf{Stage Two: object-centric reconstruction.}  
In practical applications, bounding boxes are not required in the generated video.
Moreover, in the first stage, with explicit bounding box constraints, the model learns object entities and poses, but the generated object lacks appearance details.
Therefore, in the second stage, we exclude bounding box generation and remove 3D bounding box supervision, finetuning the model to refine the object's appearance.

There are several ways to inject auxiliary 3D information, \eg, inputting depth maps or 3D bounding boxes as conditions.
Our injection-by-reconstruction approach offers two key advantages. First, the bounding boxes explicitly enhance continuous 3D awareness by serving as a pixel-level supervision target (see evaluation and supplementary material). Second, these supervisions can be easily removed by simply changing the reconstruction target, thereby mitigating inference-stage mismatches caused by inaccurate estimation of these additional signals.

\subsection{Camera-disentangled Finetuning}
\label{section:finetuning}
After the two-stage pre-training, the model learns to perceive potential rotations while preserving object shape and detailed appearance along a trajectory.
However, adapting the pre-trained model to real-world videos remains a key challenge. Unlike the synthetic dataset, where the camera is static, real-world videos often exhibit unpredictable and irregular camera motions.
Undistinguished camera motion and object motion can lead to errors in object tracking.
Therefore, we introduce camera-disentangled finetuning, which incorporates additional camera motion information into the model to help distinguish active object movement from passive camera motion (orange arrows in \cref{fig:pipeline}).

Specifically, we annotate the real-world video dataset - VIPSeg~\cite{miao2022large}, with motion trajectories and per-frame camera poses (\ie, extrinsic parameters, denoted as $\{{Cam}^i\}_{i=1}^L$). %
We incorporate camera poses as additional inputs by passing them through an MLP layer and then concatenating them with the trajectory features. The concatenated feature is then fed into a zero-initialized MLP projection layer and further input to the Traj-ControlNet.  
At inference time, it is hard to provide the camera pose information, we then randomly (50\%) drop camera poses in this fine-tuning stage, enabling the model to generate videos without camera information.
This approach helps the model differentiate between active object movement and passive camera motion, thereby improving trajectory-following accuracy and enhancing robustness across diverse video contexts.

\subsection{Training and Inference}

The mean squared error (MSE) loss is commonly used in video generation training,
and the objective is to minimize the discrepancy between the predicted noise from $\epsilon_\theta$ and the ground truth noise $\epsilon$ for each latent noise $x_t$ at timestep $t$. 
Both the two pre-training stages and fine-tuning stage leverage the same MSE loss, while the difference lies in the condition \(C^i\):

\begin{equation}
    \mathcal{L}_{\text{MSE}} = \mathbb{E}_{x_t, \epsilon} \left[ \sum_{i=1}^{L} \left\| \epsilon - \epsilon_\theta(x_t, t, C^i) \right\|^2_{2} \right],
    \label{eq1}
\end{equation}
where \(C^i\) is defined as:
\begin{equation}
    C^i =  
    \begin{cases}  
        \{I_{tr}^i, \mathbf{I}_{bbox}\}, & \text{stage one}, \\  
        \{I_{tr}^i, \mathbf{I}\}, & \text{stage two},  \\
        \{I_{tr}^i, \mathbf{I}, {Cam}^i\}, & \text{finetuning}.
    \end{cases}
    \label{eq_case}
\end{equation}
Remeber that, in the first pre-training stage, we used the bbox-augmented video frames as the reconstruction target.

\noindent\textbf{Spatial enhancement loss.}
Simply aggregating errors across all frames overlooks the spatial reconstruction accuracy of individual frames, potentially leading to object entity collapse under large rotational motions. Since each short per-frame trajectory segment provides precise localization information within individual frames, we introduce an additional spatial enhancement loss to improve spatial consistency by reconstructing the per-frame image. 
Specifically, we randomly sample the trajectory of the $j$-th frame (\ie, $I_{tr}^j$) and use it as the Traj-ControlNet condition. From its corresponding latent noise $x_{t,j}$ and the initial frame $\mathbf{I}$ ($\mathbf{I}_{bbox}$ for the first stage), the objective is to train the model to generate the target noise $\epsilon_j$ for this specific frame (\ie, single frame image reconstruction), formulated as:
\begin{equation}
   \mathcal{L}_{\text{SPA}} = {||\epsilon_j - \epsilon_\theta(x_{t,j}, t, C^j))||^2_2}, j \in (1 \sim L).
    \label{eq2}
\end{equation}
During backpropagation, only the spatial layers are updated, ensuring the refinement focuses on per-frame spatial accuracy.

Overall, the training loss function of our framework for each stage is defined as:
\begin{equation}
\mathcal{L}_\text{all} = \mathcal{L}_\text{MSE} + \lambda_\text{SPA} \mathcal{L}_\text{SPA},
\end{equation}
where $\lambda_\text{SPA}$ is a hyperparameter. 

\begin{figure*}[!ht]
\centering
    \includegraphics[width=\textwidth]{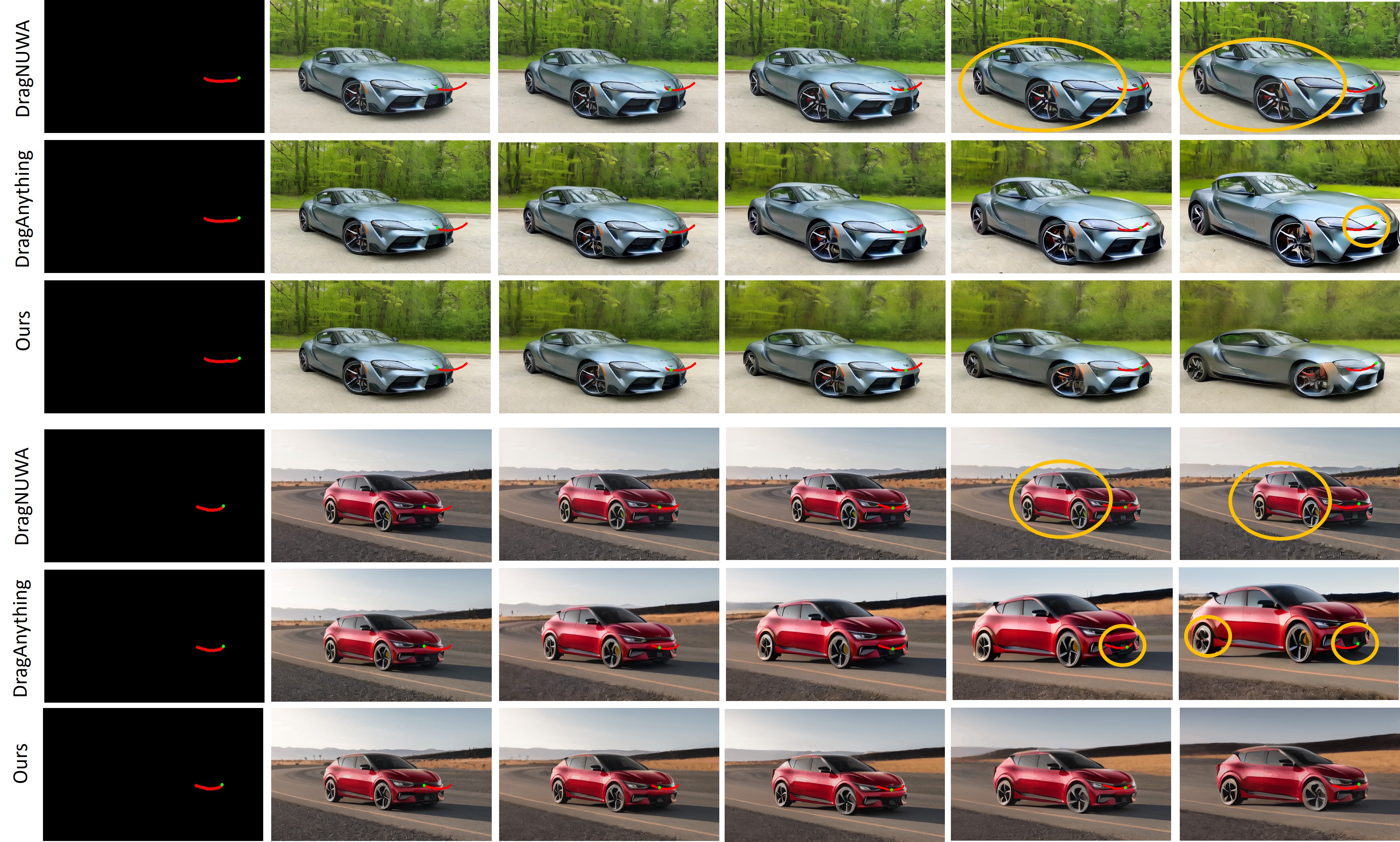}
    \vspace{-6mm}
    \caption{Visual comparison. Given the same trajectory and the initial image, our method produces plausible video frames containing the moving object following the given trajectory. In contrast, DrafNUWA and DragAnything either introduce unexpected camera motions or fail to maintain the object entity, causing severe collapse.
    }
    \label{main_res}
    \vspace{-1mm}
\end{figure*}

\begin{table*}[!ht]
\centering
\caption{Quantitative comparison between our approach and the SoTA methods.%
}
\vspace{-2mm}
\renewcommand{\arraystretch}{1.4}
\resizebox{0.8\linewidth}{!}{
\begin{tabular}{cccccccccc}
\hline
\multirow{2}{*}{\textbf{Methods}} & \multicolumn{3}{c}{\textbf{VIPSeg 256 $\times$ 256}} & \multicolumn{3}{c}{\textbf{VIPSeg 320 $\times$ 576}} & \multicolumn{3}{c}{\textbf{DAVIS 320 $\times$ 576}} \\ \cline{2-10} 
                                  & ObjMC $\downarrow$ & FID $\downarrow$ & FVD $\downarrow$ & ObjMC $\downarrow$ & FID $\downarrow$ & FVD $\downarrow$ & ObjMC $\downarrow$ & FID $\downarrow$ & FVD $\downarrow$ \\ \hline
DragNUWA 1.5 \cite{yin2023dragnuwa}       & 173.61          & 63.39           & 449.59           & 133.05          & 41.88           & 289.15    & 74.07          & 54.10          & 952.87       \\
DragAnything \cite{wu2024draganything}    & 100.23          & 61.69           & 410.70           & 91.12           & 39.29           & 275.93      &  47.01           & \textbf{50.83}           & 771.78      \\
Ours                                      & \textbf{87.56}  & \textbf{46.60}  & \textbf{384.41}  & \textbf{77.48}  & \textbf{38.41}  & \textbf{267.33} & \textbf{29.92}  & 51.48  & \textbf{729.16} \\ \hline
\end{tabular}
}
\vspace{-1mm}
\label{main_metric}
\end{table*}

\vspace{1mm}
\parag{Inference time video generation.}
During inference, given an input image, users freely draw a custom trajectory, which can be either rotational or translational, our model will generate corresponding pose-aware videos. 
Note that our approach supports more than one trajectory, see \cref{fig:intro_demo,davis_res} for examples. 
In addition, users can control the camera motion behavior by providing camera poses extracted from existing videos.

\section{Experiments and Evaluations}

\noindent \textbf{Implementation details.}  
Our base model is SVD~\cite{blattmann2023stable}, trained to generate 14 frames given an initial frame at a resolution of $320 \times 576$ pixels. All experiments are conducted using a single Tesla A100 GPU with a batch size of 1. More implementation details are in the supplementary.

\noindent\textbf{Synthetic dataset.}
For our synthetic dataset used in training, we generated 10,000 videos by sampling 5 random trajectories for each of 2,000 high-quality objects and rendering them in Blender through the Cycles engine at 5 fps. For evaluation, we create an additional trajectory for a subset of 200 sampled objects, which serves as our test set. During trajectory annotation, we augment the trajectory placement through a trajectory sampler (see the supplementary). Both training and evaluation on the synthetic dataset are conducted in the resolution of $320\times576$ pixels with 14 frames per sequence.

\noindent\textbf{Real-world dataset.}  
For real-world finetuning, we adopted the VIPSeg dataset~\cite{miao2022large}, following DragAnything. VIPSeg is a large-scale video segmentation dataset comprising over 3,000 high-quality videos with annotated object masks. We utilized its training set for finetuning and its validation set for evaluation, employing the same trajectory extraction pipeline as in DragAnything (see supplementary).  

To further evaluate the generalization capability of our approach, we additionally select the DAVIS dataset~\cite{perazzi2016benchmark}, a high-quality video segmentation benchmark, for out-of-distribution (OOD) evaluation. Note that, we only annotate the trajectory for this dataset for OOD evaluation, and at inference, camera poses are not used.

\noindent\textbf{Evaluation metrics.}
We follow established evaluation metrics from previous trajectory-guided studies~\cite{pan2023draggan, yin2023dragnuwa, wu2024draganything}, assessing performance across two key aspects: 
\begin{itemize}
    \item \textbf{Controlling accuracy}. To quantify the precision of trajectory-guided dragging, we use the ObjMC metric, which computes the mean squared error between the temporal motion tendency of generated samples and that of the ground truth videos.  
    \item \textbf{Generation quality}. We employ the Fréchet Inception Distance (FID)~\cite{heusel2017fid} to assess the single-frame spatial quality and the Fréchet Video Distance (FVD)~\cite{unterthiner2018fvd} to measure temporal consistency.
\end{itemize}

\subsection{Comparison}

\noindent\textbf{Competitors.} We primarily evaluate our model's performance against two recent trajectory-guided video generation models: DragNUWA 1.5~\cite{yin2023dragnuwa} and DragAnything~\cite{wu2024draganything}.  
DragNUWA is the first method designed for open-domain video dragging. It utilizes optical flow estimated by UniMatch~\cite{unimatch} as trajectory guidance, combining these trajectories with text and image conditions to enable motion-controllable fine-tuning. Built upon DragNUWA, DragAnything enhances object-level animation by extracting entity feature maps from instance masks and integrating them into ControlNet for further fine-tuning.  
To ensure fair comparisons, the same trajectory is converted into each model’s respective trajectory representation (\eg, the trajectory point set or trajectory segment images).
Additionally, since DragAnything reports metric performance only at a resolution of $256 \times 256$, we evaluate our model at both the original $320 \times 576$ training resolution and the downsampled $256 \times 256$ resolution to ensure a fair comparison.
Due to the unavailability of public model weights for Tora and MotionCtrl with SVD pretraining, we exclude these models from our comparison.

\noindent \textbf{Quantitative Results.} 
As shown in \cref{main_metric}, \shortname outperforms competing methods across both resolutions in all trajectory-specific and video quality metrics, with particularly strong performance in lower-resolution synthesis.
In terms of trajectory accuracy, our model achieves a significant reduction in error than DragAnything, with improvements of 13\%, 15\%, and 36\% for lower and standard resolutions on the VIPSeg dataset and standard resolution on the DAVIS dataset, respectively. These results demonstrate the effectiveness of our pose-aware pretraining approach, which leverages high-precision synthetic data pairs with 3D bounding boxes and incorporates a camera-disentanglement module to enhance the model’s understanding of object-oriented motion and passive camera movement.

For single-frame reconstruction performance, measured by FID, our model outperforms DragAnything by a significant margin, achieving an improvement of $\textbf{15.01}$ at the downsampled $256 \times 256$ resolution and 0.88 at the standard $320 \times 576$ resolution on the VIPSeg dataset.
Furthermore, our model demonstrates improved temporal consistency across frames, achieving substantial reductions in FVD scores: 26.29, 8.6, and 42.62 across the three evaluation settings. These results highlight our model's significant advancements in trajectory-guided motion control, achieving precise dragging-following and reliable motion. Notably, our approach shows strong performance in traditional in-distribution scenarios while exhibiting even greater superiority in out-of-distribution (OOD) settings (see the evaluation on DAVIS in \cref{main_metric}). 

\begin{figure}[!t]
\centering
{
\includegraphics[width=1\columnwidth]{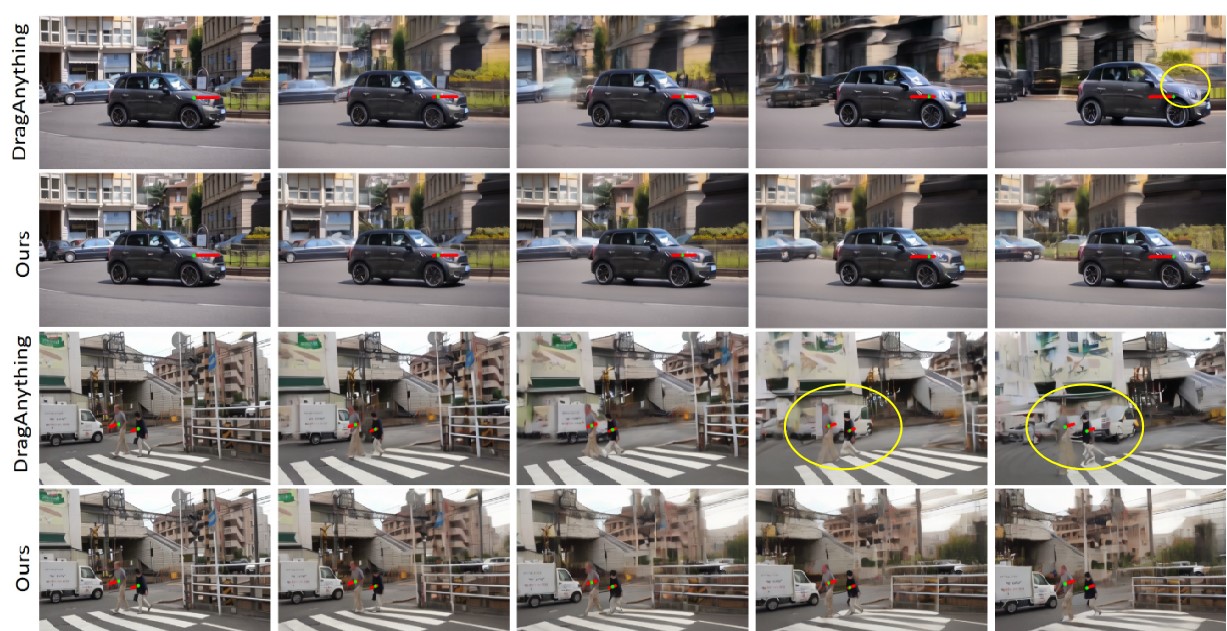}} 
\vspace{-6mm}
\caption{Transitional samples on DAVIS dataset.}
\vspace{-4mm}
\label{davis_res}
\end{figure}

\noindent \textbf{Qualitative Results.}
Visual comparisons are displayed in \cref{main_res}, where \shortname achieves superior performance in generating 3D and pose-aligned rotational motions that accurately follow 2D trajectories with precise pixel-level trajectory adherence.  
Furthermore, our model demonstrates enhanced capability in reliably reconstructing occluded object regions, even under changing 6D poses, such as rotational car doors.
In contrast, DragNUWA primarily generates moving camera perspectives while keeping object poses static, resulting in poor rotational alignment. 
DragAnything produces unstable and mismatched rotational motions, where dragged pixels progressively drift across frames, potentially leading to entity collapse.
This comparison highlights the effectiveness of our approach in generating more accurate, stable, and realistic motion sequences, particularly when handling complex trajectories. 

Additionally, in \cref{davis_res}, we show visual results on OOD examples, where the trajectory is mostly translational instead of rotational. Our generated videos are superior to DragAnything.

\begin{figure}[!h]
\centering
{
\includegraphics[width=0.9\columnwidth]{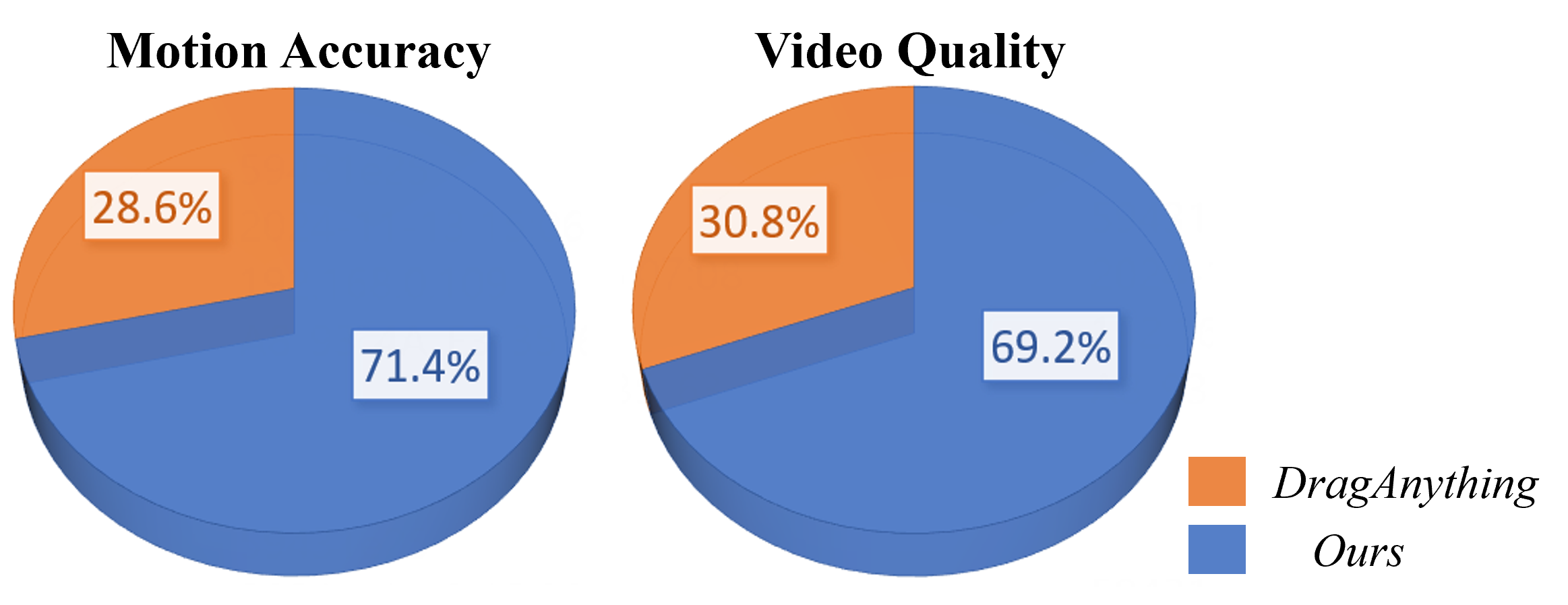}}
\vspace{-2mm}
\caption{
The voting results of the user evaluation.}
\label{userstudy}
\vspace{-0.3cm}
\end{figure}

\noindent \textbf{User Evaluation.} 
Other than the visual and statistical evaluations, we conduct a subjective user evaluation, where participants rate generated videos (20 sampled videos from the VIPSeg test set and manually annotated wild images) based on two criteria: overall motion accuracy and visual quality. 
We have invited seven experienced experts on the video generation topic and asked them to rate the videos.  

As shown in Fig.~\ref{userstudy}, our approach is favorably voted, receiving \textbf{43\%} more votes in trajectory-following accuracy and \textbf{39\%} more votes in video quality.
These findings are consistent with the superior performance observed in quantitative metrics and qualitative visualizations.

\begin{table}[!t]
\centering
\caption{Ablation results on the real-world video test set with several model variants.}
\vspace{-2mm}
\renewcommand{\arraystretch}{1.1}
\resizebox{0.65\linewidth}{!}{
\begin{tabular}{cccc}
\hline
\multicolumn{1}{c}{\multirow{2}{*}{\textbf{Methods}}} & \multicolumn{3}{c}{320 $\times$ 576} \\ \cline{2-4}
\multicolumn{1}{c}{}                                &ObjMC $\downarrow$ &          FID $\downarrow$ & FVD $\downarrow$ \\ \hline
Full method                                          & 77.48                   & 38.41                  & 267.33           \\ \hline
No bbox stage                                       & 81.36         & 41.90           & 275.40           \\ 
No pretrain                                   & 145.72        & 42.62          & 486.84          \\ \hline
No Cam-disen                                          & 83.22         & 39.71        & 
279.15          \\ 
No SPA-loss                                   & 137.26        & 39.79          & 436.56          \\
\hline
\end{tabular}
}
\vspace{-0.2cm}
\label{ablation}
\end{table}

\begin{table}[!t]
\centering
\caption{Comparison results on the synthetic validation set of our designed ablation studies on the pretraining stage. 
}
\vspace{-2mm}
\renewcommand{\arraystretch}{1.1}
\resizebox{0.7\linewidth}{!}{
\begin{tabular}{cccc}
\hline
\multicolumn{1}{c}{\multirow{2}{*}{\textbf{Methods}}} & \multicolumn{3}{c}{Synthetic Dataset Val} \\ \cline{2-4}
\multicolumn{1}{c}{}                                &ObjMC $\downarrow$ &          FID $\downarrow$ & FVD $\downarrow$ \\ \hline
Full method                                          & 0.196         & 47.18                & 187.12           \\ \hline
No bbox stage                                       & 0.266         & 50.30           & 223.74           \\ 
No SPA-loss                                   & 0.265        & 46.72          & 197.68          \\
\hline
\end{tabular}
}
\vspace{-0.3cm}
\label{ablation2}
\end{table}

\subsection{Ablation Study}

\noindent \textbf{Pose-aware pretraining.} 
To evaluate the impact of our pretraining strategy, we propose two experimental variants: (1) training the model without 3D-aware pertaining (denoted as `No pretrain'), and (2) training the model using only the second-stage pretraining while omitting 3D bounding box supervision (denoted as `No bbox stage').

\begin{figure}[!t]
\centering
{
\includegraphics[width=1\columnwidth]{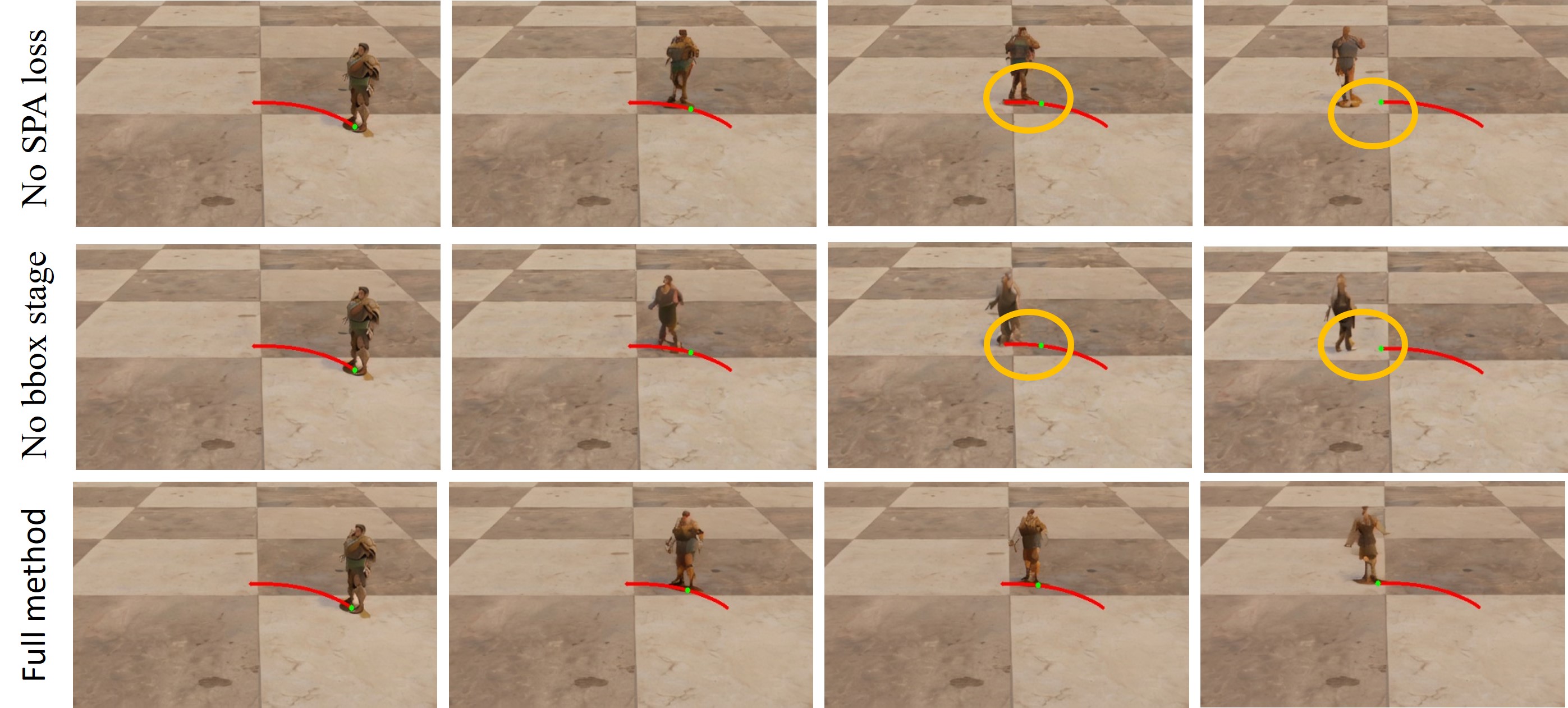}}
\vspace{-6mm}
\caption{
Visualization of ablation results in synthetic dataset. %
}
\label{ab_blender}
\vspace{-0.6cm}
\end{figure}

As shown in \cref{ablation,ablation2}, excluding 3D bounding box supervision during pretraining leads to a degradation in performance across all evaluation metrics on both the synthetic and open-domain datasets. 
Notably, in \cref{ablation}, compared with `No pretraining', the absence of 3D bounding box pre-training has minimal impact on the metric. 
This is because 3D bounding boxes mainly affect object pose localization, and inaccurate poses do not significantly degrade motion accuracy or quality. 
However, the visual results in \cref{ab_blender} clearly demonstrate the defects without bounding box pre-training, \ie, there is pronounced object collapse, exacerbating trajectory misalignment. More visual comparisons can be found in the supplementary material.
Furthermore, omitting the two-stage pretraining approach results in even greater performance degradation, leading to a substantial drop in trajectory-following capability.

\noindent \textbf{Spatial enhancement loss.}
Compared to standard video generation tasks, achieving both precise trajectory control and spatial visual consistency is significantly more challenging, especially when the motion involves complex dynamics or large-angle rotations. Consequently, integrating spatial enhancement loss and formulating image-to-image prediction as a sub-task during training becomes essential. As demonstrated in \cref{ablation,ablation2,ab_blender}, models trained without spatial loss (denoted as `No SPA-loss') show considerable degradation in trajectory alignment reliability and generation quality, with frequent blurring observed for fast-moving objects.

\noindent \textbf{Camera disentanglement.}
Considering the unexpected influence of camera motions, we incorporate camera disentangled fine-tuning to enhance the model's trajectory, controlling awareness of generating naturally activated object motion with passive camera movements. 
We evaluate the impact of this design by discarding camera poses (denoted as `No Cam-disen').  Experimental results (\cref{ablation}) indicate that models without camera disentangling exhibit lower object motion-matching accuracy and a corresponding decrease in the FVD metric. This inaccuracy arises from the model's uncertainty regarding how to generate aligned passive camera motions and results in a reduced ability to handle trajectory variations involving background motion.

We show more visual results for ablation experiments in the 
supplementary.

\section{Conclusion}

In this paper, we have presented \shortname, a novel approach for pose-aware trajectory control in video generation. We propose a two-stage pre-training strategy to enable the accurate trajectory-following ability, while explicitly exploiting 3D bounding boxes as a supervision signal to perceive the object's potential 6D pose changes along the driving trajectory. A new dataset, \shortdname, is constructed to achieve our goal. Our approach outperforms state-of-the-art methods quantitatively and qualitatively. We hope that the 3D bounding box signal and the injection-by-reconstruction design can inspire future research in controllable video generation or relevant topics.

\parag{Acknowledgment.} LB completed part of this work during his Master's study at the University of Edinburgh. CJ was supported by a gift from Adobe.

{
    \small
    \bibliographystyle{ieeenat_fullname}
    \bibliography{main}

\begin{thebibliography}{48}
\providecommand{\natexlab}[1]{#1}
\providecommand{\url}[1]{\texttt{#1}}
\expandafter\ifx\csname urlstyle\endcsname\relax
  \providecommand{\doi}[1]{doi: #1}\else
  \providecommand{\doi}{doi: \begingroup \urlstyle{rm}\Url}\fi

\bibitem[Blattmann et~al.(2021)Blattmann, Milbich, Dorkenwald, and Ommer]{blattmann2021ipoke}
Andreas Blattmann, Timo Milbich, Michael Dorkenwald, and Bj{\"o}rn Ommer.
\newblock ipoke: Poking a still image for controlled stochastic video synthesis.
\newblock In \emph{Proceedings of the IEEE/CVF International Conference on Computer Vision}, pages 14707--14717, 2021.

\bibitem[Blattmann et~al.(2023{\natexlab{a}})Blattmann, Dockhorn, Kulal, Mendelevitch, Kilian, Lorenz, Levi, English, Voleti, Letts, et~al.]{blattmann2023stable}
Andreas Blattmann, Tim Dockhorn, Sumith Kulal, Daniel Mendelevitch, Maciej Kilian, Dominik Lorenz, Yam Levi, Zion English, Vikram Voleti, Adam Letts, et~al.
\newblock Stable video diffusion: Scaling latent video diffusion models to large datasets.
\newblock \emph{arXiv preprint arXiv:2311.15127}, 2023{\natexlab{a}}.

\bibitem[Blattmann et~al.(2023{\natexlab{b}})Blattmann, Rombach, Ling, Dockhorn, Kim, Fidler, and Kreis]{blattmann2023videoldm}
Andreas Blattmann, Robin Rombach, Huan Ling, Tim Dockhorn, Seung~Wook Kim, Sanja Fidler, and Karsten Kreis.
\newblock Align your latents: High-resolution video synthesis with latent diffusion models.
\newblock In \emph{Proceedings of the IEEE/CVF Conference on Computer Vision and Pattern Recognition}, pages 22563--22575, 2023{\natexlab{b}}.

\bibitem[Brooks et~al.(2024)Brooks, Peebles, Holmes, DePue, Guo, Jing, Schnurr, Taylor, Luhman, Luhman, et~al.]{brooks2024video}
Tim Brooks, Bill Peebles, Connor Holmes, Will DePue, Yufei Guo, Li Jing, David Schnurr, Joe Taylor, Troy Luhman, Eric Luhman, et~al.
\newblock Video generation models as world simulators. 2024.
\newblock \emph{URL https://openai. com/research/video-generation-models-as-world-simulators}, 3, 2024.

\bibitem[Caraffa et~al.(2024)Caraffa, Boscaini, Hamza, and Poiesi]{caraffa2024freeze}
Andrea Caraffa, Davide Boscaini, Amir Hamza, and Fabio Poiesi.
\newblock Freeze: Training-free zero-shot 6d pose estimation with geometric and vision foundation models.
\newblock \emph{European Conference on Computer Vision (ECCV)}, 2024.

\bibitem[Chen et~al.(2023{\natexlab{a}})Chen, Xia, He, Zhang, Cun, Yang, Xing, Liu, Chen, Wang, Weng, and Shan]{chen2023videocrafter1}
Haoxin Chen, Menghan Xia, Yingqing He, Yong Zhang, Xiaodong Cun, Shaoshu Yang, Jinbo Xing, Yaofang Liu, Qifeng Chen, Xintao Wang, Chao Weng, and Ying Shan.
\newblock Videocrafter1: Open diffusion models for high-quality video generation, 2023{\natexlab{a}}.

\bibitem[Chen et~al.(2023{\natexlab{b}})Chen, Lin, Tseng, Lin, and Yang]{chen2023motion}
Tsai-Shien Chen, Chieh~Hubert Lin, Hung-Yu Tseng, Tsung-Yi Lin, and Ming-Hsuan Yang.
\newblock Motion-conditioned diffusion model for controllable video synthesis.
\newblock \emph{arXiv preprint arXiv:2304.14404}, 2023{\natexlab{b}}.

\bibitem[Dang et~al.(2024)Dang, Huang, Wang, and Salzmann]{dang2024openmaterial}
Zheng Dang, Jialu Huang, Fei Wang, and Mathieu Salzmann.
\newblock Openmaterial: A comprehensive dataset of complex materials for 3d reconstruction.
\newblock \emph{arXiv preprint arXiv:2406.08894}, 2024.

\bibitem[Deitke et~al.(2024)Deitke, Liu, Wallingford, Ngo, Michel, Kusupati, Fan, Laforte, Voleti, Gadre, et~al.]{deitke2024objaverse}
Matt Deitke, Ruoshi Liu, Matthew Wallingford, Huong Ngo, Oscar Michel, Aditya Kusupati, Alan Fan, Christian Laforte, Vikram Voleti, Samir~Yitzhak Gadre, et~al.
\newblock Objaverse-xl: A universe of 10m+ 3d objects.
\newblock \emph{Advances in Neural Information Processing Systems}, 36, 2024.

\bibitem[Esser et~al.(2023)Esser, Chiu, Atighehchian, Granskog, and Germanidis]{esser2023structure}
Patrick Esser, Johnathan Chiu, Parmida Atighehchian, Jonathan Granskog, and Anastasis Germanidis.
\newblock Structure and content-guided video synthesis with diffusion models.
\newblock In \emph{Proceedings of the IEEE/CVF International Conference on Computer Vision}, pages 7346--7356, 2023.

\bibitem[He et~al.(2022)He, Yang, Zhang, Shan, and Chen]{he2022lvdm}
Yingqing He, Tianyu Yang, Yong Zhang, Ying Shan, and Qifeng Chen.
\newblock Latent video diffusion models for high-fidelity long video generation.
\newblock \emph{arXiv preprint arXiv:2211.13221}, 2022.

\bibitem[Heusel et~al.(2017)Heusel, Ramsauer, Unterthiner, Nessler, and Hochreiter]{heusel2017fid}
Martin Heusel, Hubert Ramsauer, Thomas Unterthiner, Bernhard Nessler, and Sepp Hochreiter.
\newblock Gans trained by a two time-scale update rule converge to a local nash equilibrium.
\newblock \emph{Advances in neural information processing systems}, 30, 2017.

\bibitem[Hong et~al.(2022)Hong, Ding, Zheng, Liu, and Tang]{hong2022cogvideo}
Wenyi Hong, Ming Ding, Wendi Zheng, Xinghan Liu, and Jie Tang.
\newblock Cogvideo: Large-scale pretraining for text-to-video generation via transformers.
\newblock \emph{arXiv preprint arXiv:2205.15868}, 2022.

\bibitem[Hu(2024)]{hu2024animate}
Li Hu.
\newblock Animate anyone: Consistent and controllable image-to-video synthesis for character animation.
\newblock In \emph{Proceedings of the IEEE/CVF Conference on Computer Vision and Pattern Recognition}, pages 8153--8163, 2024.

\bibitem[Karaev et~al.(2024)Karaev, Rocco, Graham, Neverova, Vedaldi, and Rupprecht]{karaev23cotracker}
Nikita Karaev, Ignacio Rocco, Benjamin Graham, Natalia Neverova, Andrea Vedaldi, and Christian Rupprecht.
\newblock Cotracker: It is better to track together.
\newblock In \emph{Proc. {ECCV}}, 2024.

\bibitem[Kirillov et~al.(2023)Kirillov, Mintun, Ravi, Mao, Rolland, Gustafson, Xiao, Whitehead, Berg, Lo, et~al.]{kirillov2023segment}
Alexander Kirillov, Eric Mintun, Nikhila Ravi, Hanzi Mao, Chloe Rolland, Laura Gustafson, Tete Xiao, Spencer Whitehead, Alexander~C Berg, Wan-Yen Lo, et~al.
\newblock Segment anything.
\newblock In \emph{Proceedings of the IEEE/CVF International Conference on Computer Vision}, pages 4015--4026, 2023.

\bibitem[Li et~al.(2024)Li, Zheng, Rupprecht, and Vedaldi]{li2024puppet}
Ruining Li, Chuanxia Zheng, Christian Rupprecht, and Andrea Vedaldi.
\newblock Puppet-master: Scaling interactive video generation as a motion prior for part-level dynamics.
\newblock \emph{arXiv preprint arXiv:2408.04631}, 2024.

\bibitem[Ma et~al.(2024)Ma, He, Cun, Wang, Chen, Li, and Chen]{ma2024follow}
Yue Ma, Yingqing He, Xiaodong Cun, Xintao Wang, Siran Chen, Xiu Li, and Qifeng Chen.
\newblock Follow your pose: Pose-guided text-to-video generation using pose-free videos.
\newblock In \emph{Proceedings of the AAAI Conference on Artificial Intelligence}, pages 4117--4125, 2024.

\bibitem[Miao et~al.(2022)Miao, Wang, Wu, Li, Zhang, Wei, and Yang]{miao2022large}
Jiaxu Miao, Xiaohan Wang, Yu Wu, Wei Li, Xu Zhang, Yunchao Wei, and Yi Yang.
\newblock Large-scale video panoptic segmentation in the wild: A benchmark.
\newblock In \emph{Proceedings of the {IEEE} Conference on Computer Vision and Pattern Recognition}, 2022.

\bibitem[Ni et~al.(2023)Ni, Shi, Li, Huang, and Min]{ni2023conditional}
Haomiao Ni, Changhao Shi, Kai Li, Sharon~X Huang, and Martin~Renqiang Min.
\newblock Conditional image-to-video generation with latent flow diffusion models.
\newblock In \emph{Proceedings of the IEEE/CVF conference on computer vision and pattern recognition}, pages 18444--18455, 2023.

\bibitem[Ni et~al.(2024)Ni, Egger, Lohit, Cherian, Wang, Koike-Akino, Huang, and Marks]{ni2024ti2v}
Haomiao Ni, Bernhard Egger, Suhas Lohit, Anoop Cherian, Ye Wang, Toshiaki Koike-Akino, Sharon~X Huang, and Tim~K Marks.
\newblock Ti2v-zero: Zero-shot image conditioning for text-to-video diffusion models.
\newblock In \emph{Proceedings of the IEEE/CVF Conference on Computer Vision and Pattern Recognition}, pages 9015--9025, 2024.

\bibitem[{OpenAI}(2024)]{GPT:24}
{OpenAI}.
\newblock {GPT (v4) [Large language model]}.
\newblock \url{https://chat.openai.com}, 2024.

\bibitem[{\"O}rnek et~al.(2025){\"O}rnek, Labb{\'e}, Tekin, Ma, Keskin, Forster, and Hodan]{ornek2025foundpose}
Evin~P{\i}nar {\"O}rnek, Yann Labb{\'e}, Bugra Tekin, Lingni Ma, Cem Keskin, Christian Forster, and Tomas Hodan.
\newblock Foundpose: Unseen object pose estimation with foundation features.
\newblock In \emph{European Conference on Computer Vision}, pages 163--182. Springer, 2025.

\bibitem[Pan et~al.(2023)Pan, Tewari, Leimk{\"u}hler, Liu, Meka, and Theobalt]{pan2023draggan}
Xingang Pan, Ayush Tewari, Thomas Leimk{\"u}hler, Lingjie Liu, Abhimitra Meka, and Christian Theobalt.
\newblock Drag your gan: Interactive point-based manipulation on the generative image manifold.
\newblock In \emph{ACM SIGGRAPH 2023 Conference Proceedings}, pages 1--11, 2023.

\bibitem[Perazzi et~al.(2016)Perazzi, Pont-Tuset, McWilliams, Van~Gool, Gross, and Sorkine-Hornung]{perazzi2016benchmark}
Federico Perazzi, Jordi Pont-Tuset, Brian McWilliams, Luc Van~Gool, Markus Gross, and Alexander Sorkine-Hornung.
\newblock A benchmark dataset and evaluation methodology for video object segmentation.
\newblock In \emph{Proceedings of the IEEE conference on computer vision and pattern recognition}, pages 724--732, 2016.

\bibitem[Ramesh et~al.(2022)Ramesh, Dhariwal, Nichol, Chu, and Chen]{ramesh2022hierarchical}
Aditya Ramesh, Prafulla Dhariwal, Alex Nichol, Casey Chu, and Mark Chen.
\newblock Hierarchical text-conditional image generation with clip latents.
\newblock \emph{arXiv preprint arXiv:2204.06125}, 1\penalty0 (2):\penalty0 3, 2022.

\bibitem[Rombach et~al.(2022)Rombach, Blattmann, Lorenz, Esser, and Ommer]{rombach2022high}
Robin Rombach, Andreas Blattmann, Dominik Lorenz, Patrick Esser, and Bj{\"o}rn Ommer.
\newblock High-resolution image synthesis with latent diffusion models.
\newblock In \emph{Proceedings of the IEEE/CVF conference on computer vision and pattern recognition}, pages 10684--10695, 2022.

\bibitem[Ronneberger et~al.(2015)Ronneberger, Fischer, and Brox]{ronneberger2015u}
Olaf Ronneberger, Philipp Fischer, and Thomas Brox.
\newblock U-net: Convolutional networks for biomedical image segmentation.
\newblock In \emph{Medical image computing and computer-assisted intervention--MICCAI 2015: 18th international conference, Munich, Germany, October 5-9, 2015, proceedings, part III 18}, pages 234--241. Springer, 2015.

\bibitem[Saharia et~al.(2022)Saharia, Chan, Saxena, Li, Whang, Denton, Ghasemipour, Gontijo~Lopes, Karagol~Ayan, Salimans, et~al.]{saharia2022photorealistic}
Chitwan Saharia, William Chan, Saurabh Saxena, Lala Li, Jay Whang, Emily~L Denton, Kamyar Ghasemipour, Raphael Gontijo~Lopes, Burcu Karagol~Ayan, Tim Salimans, et~al.
\newblock Photorealistic text-to-image diffusion models with deep language understanding.
\newblock \emph{Advances in neural information processing systems}, 35:\penalty0 36479--36494, 2022.

\bibitem[Singer et~al.(2022)Singer, Polyak, Hayes, Yin, An, Zhang, Hu, Yang, Ashual, Gafni, et~al.]{singer2022make}
Uriel Singer, Adam Polyak, Thomas Hayes, Xi Yin, Jie An, Songyang Zhang, Qiyuan Hu, Harry Yang, Oron Ashual, Oran Gafni, et~al.
\newblock Make-a-video: Text-to-video generation without text-video data.
\newblock \emph{arXiv preprint arXiv:2209.14792}, 2022.

\bibitem[Teed and Deng(2021)]{teed2021droid}
Zachary Teed and Jia Deng.
\newblock {DROID-SLAM: Deep Visual SLAM for Monocular, Stereo, and RGB-D Cameras}.
\newblock \emph{Advances in neural information processing systems}, 2021.

\bibitem[Tian et~al.(2024)Tian, Wang, Zhang, and Bo]{tian2024emo}
Linrui Tian, Qi Wang, Bang Zhang, and Liefeng Bo.
\newblock Emo: Emote portrait alive - generating expressive portrait videos with audio2video diffusion model under weak conditions, 2024.

\bibitem[Unterthiner et~al.(2018)Unterthiner, Van~Steenkiste, Kurach, Marinier, Michalski, and Gelly]{unterthiner2018fvd}
Thomas Unterthiner, Sjoerd Van~Steenkiste, Karol Kurach, Raphael Marinier, Marcin Michalski, and Sylvain Gelly.
\newblock Towards accurate generative models of video: A new metric \& challenges.
\newblock \emph{arXiv preprint arXiv:1812.01717}, 2018.

\bibitem[Wang et~al.(2024{\natexlab{a}})Wang, Zhang, Zou, Zeng, Wei, Yuan, and Li]{wang2024boximator}
Jiawei Wang, Yuchen Zhang, Jiaxin Zou, Yan Zeng, Guoqiang Wei, Liping Yuan, and Hang Li.
\newblock Boximator: Generating rich and controllable motions for video synthesis, 2024{\natexlab{a}}.

\bibitem[Wang et~al.(2024{\natexlab{b}})Wang, Zhang, Gao, Wang, Zhou, Zhang, Yan, and Sang]{wang2024unianimate}
Xiang Wang, Shiwei Zhang, Changxin Gao, Jiayu Wang, Xiaoqiang Zhou, Yingya Zhang, Luxin Yan, and Nong Sang.
\newblock Unianimate: Taming unified video diffusion models for consistent human image animation.
\newblock \emph{arXiv preprint arXiv:2406.01188}, 2024{\natexlab{b}}.

\bibitem[Wang et~al.(2023)Wang, Yuan, Wang, Chen, Xia, Luo, and Shan]{wang2023motionctrl}
Zhouxia Wang, Ziyang Yuan, Xintao Wang, Tianshui Chen, Menghan Xia, Ping Luo, and Ying Shan.
\newblock Motionctrl: A unified and flexible motion controller for video generation.
\newblock \emph{arXiv preprint arXiv:2312.03641}, 2023.

\bibitem[Wu et~al.(2022)Wu, Liang, Ji, Yang, Fang, Jiang, and Duan]{wu2022nuwa}
Chenfei Wu, Jian Liang, Lei Ji, Fan Yang, Yuejian Fang, Daxin Jiang, and Nan Duan.
\newblock N{\"u}wa: Visual synthesis pre-training for neural visual world creation.
\newblock In \emph{European conference on computer vision}, pages 720--736. Springer, 2022.

\bibitem[Wu et~al.(2024)Wu, Li, Gu, Zhao, He, Zhang, Shou, Li, Gao, and Zhang]{wu2024draganything}
Wejia Wu, Zhuang Li, Yuchao Gu, Rui Zhao, Yefei He, David~Junhao Zhang, Mike~Zheng Shou, Yan Li, Tingting Gao, and Di Zhang.
\newblock Draganything: Motion control for anything using entity representation.
\newblock \emph{arXiv preprint arXiv:2403.07420}, 2024.

\bibitem[Xu et~al.(2024)Xu, Li, Su, Shang, Zhang, Liu, Wang, Yao, and Zhu]{xu2024hallo}
Mingwang Xu, Hui Li, Qingkun Su, Hanlin Shang, Liwei Zhang, Ce Liu, Jingdong Wang, Yao Yao, and Siyu Zhu.
\newblock Hallo: Hierarchical audio-driven visual synthesis for portrait image animation.
\newblock \emph{arXiv preprint arXiv:2406.08801}, 2024.

\bibitem[Yang et~al.(2023)Yang, Qi, Feng, Zhang, and Shi]{unimatch}
Lihe Yang, Lei Qi, Litong Feng, Wayne Zhang, and Yinghuan Shi.
\newblock Revisiting weak-to-strong consistency in semi-supervised semantic segmentation.
\newblock In \emph{CVPR}, 2023.

\bibitem[Yang et~al.(2024)Yang, Hou, Huang, Ma, Wan, Zhang, Chen, and Liao]{yang2024direct}
Shiyuan Yang, Liang Hou, Haibin Huang, Chongyang Ma, Pengfei Wan, Di Zhang, Xiaodong Chen, and Jing Liao.
\newblock Direct-a-video: Customized video generation with user-directed camera movement and object motion.
\newblock In \emph{ACM SIGGRAPH 2024 Conference Papers}, pages 1--12, 2024.

\bibitem[Yin et~al.(2023)Yin, Wu, Liang, Shi, Li, Ming, and Duan]{yin2023dragnuwa}
Shengming Yin, Chenfei Wu, Jian Liang, Jie Shi, Houqiang Li, Gong Ming, and Nan Duan.
\newblock Dragnuwa: Fine-grained control in video generation by integrating text, image, and trajectory.
\newblock \emph{arXiv preprint arXiv:2308.08089}, 2023.

\bibitem[Zeng et~al.(2024)Zeng, Deschaintre, Georgiev, Hold-Geoffroy, Hu, Luan, Yan, and Ha{\v{s}}an]{zeng2024rgb}
Zheng Zeng, Valentin Deschaintre, Iliyan Georgiev, Yannick Hold-Geoffroy, Yiwei Hu, Fujun Luan, Ling-Qi Yan, and Milo{\v{s}} Ha{\v{s}}an.
\newblock Rgbx: Image decomposition and synthesis using material-and lighting-aware diffusion models.
\newblock In \emph{ACM SIGGRAPH 2024 Conference Papers}, pages 1--11, 2024.

\bibitem[Zhang et~al.(2024{\natexlab{a}})Zhang, Wu, Cruz~Gambardella, Huang, Phung, Ouyang, and Cai]{panfusion2024}
Cheng Zhang, Qianyi Wu, Camilo Cruz~Gambardella, Xiaoshui Huang, Dinh Phung, Wanli Ouyang, and Jianfei Cai.
\newblock Taming stable diffusion for text to 360◦ panorama image generation.
\newblock In \emph{Proceedings of the IEEE/CVF Conference on Computer Vision and Pattern Recognition}, 2024{\natexlab{a}}.

\bibitem[Zhang et~al.(2024{\natexlab{b}})Zhang, Liao, Li, Qin, and Wang]{zhang2024tora}
Zhenghao Zhang, Junchao Liao, Menghao Li, Long Qin, and Weizhi Wang.
\newblock Tora: Trajectory-oriented diffusion transformer for video generation.
\newblock \emph{arXiv preprint arXiv:2407.21705}, 2024{\natexlab{b}}.

\bibitem[Zhou et~al.(2018)Zhou, Tucker, Flynn, Fyffe, and Snavely]{zhou2018stereo}
Tinghui Zhou, Richard Tucker, John Flynn, Graham Fyffe, and Noah Snavely.
\newblock Stereo magnification: Learning view synthesis using multiplane images.
\newblock \emph{arXiv preprint arXiv:1805.09817}, 2018.

\bibitem[Zhu et~al.(2022)Zhu, Luan, Huo, Lin, Zhong, Xi, Wang, Bao, Zheng, and Tang]{zhu2022learning}
Jingsen Zhu, Fujun Luan, Yuchi Huo, Zihao Lin, Zhihua Zhong, Dianbing Xi, Rui Wang, Hujun Bao, Jiaxiang Zheng, and Rui Tang.
\newblock Learning-based inverse rendering of complex indoor scenes with differentiable monte carlo raytracing.
\newblock In \emph{SIGGRAPH Asia 2022 Conference Papers}, pages 1--8, 2022.

\bibitem[Zhu et~al.(2024)Zhu, Chen, Dai, Su, Xu, Cao, Yao, Zhu, and Zhu]{zhu2024champ}
Shenhao Zhu, Junming~Leo Chen, Zuozhuo Dai, Qingkun Su, Yinghui Xu, Xun Cao, Yao Yao, Hao Zhu, and Siyu Zhu.
\newblock Champ: Controllable and consistent human image animation with 3d parametric guidance.
\newblock \emph{arXiv preprint arXiv:2403.14781}, 2024.

\end{thebibliography}
}

\clearpage
\appendix
\setcounter{page}{1}

\setcounter{table}{0}
\renewcommand{\thetable}{A\arabic{table}}
\setcounter{figure}{0}
\renewcommand{\thefigure}{A\arabic{figure}}

\maketitlesupplementary

\section{Synthetic Data Consturction}

\subsection{Construction Pipeline}
In this section, we describe the detailed pipeline for constructing synthetic data. First, we set up a realistic virtual scene featuring a fixed camera, a wood-textured floor, and indoor HDRI images to simulate natural indoor environments, including floor texture and lighting. Next, we sample an object from a filtered subset of 2,000 objects from Objaverse \cite{deitke2024objaverse}. The object is then normalized to a height of 1 unit and placed on the floor.

\begin{figure*}[t]
\centering
{
\includegraphics[width=2\columnwidth]{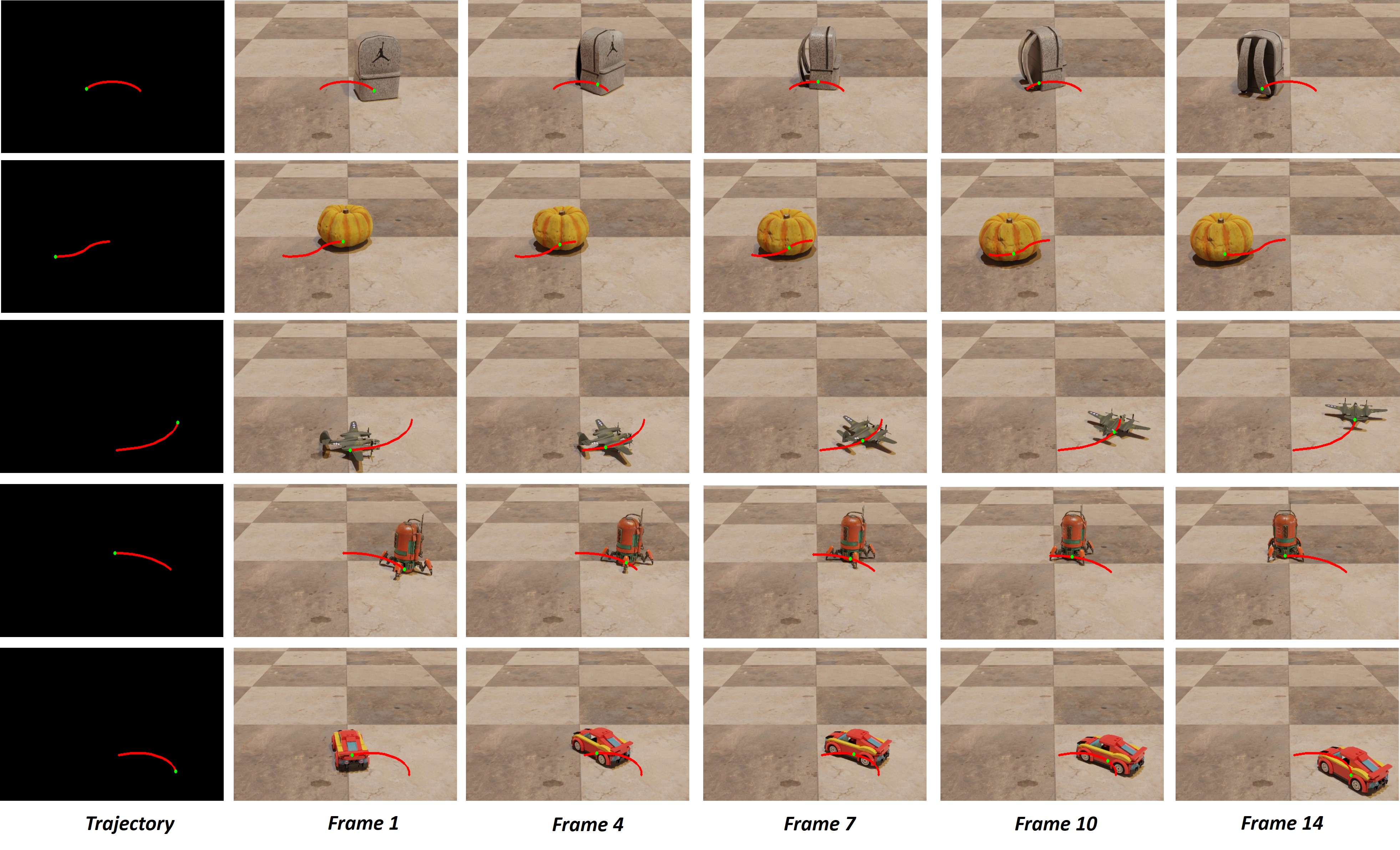} }
\vspace{-3mm}
\caption{
Visualization for several animated samples from our trajectory augmented synthetic dataset.}
\label{data_show}
\vspace{-0.2cm}
\end{figure*}

We then generate a random trajectory by defining a curve with a randomly initialized starting point, rotational angle, and length. The starting points of the trajectories are randomly sampled within a circle of radius 1 unit, centered at the origin $(0,0)$. The initial orientation of the object is set at a random angle between 0° and 90° relative to the positive x-axis. Two types of trajectory templates are defined: i) a circular trajectory without any turning points, and ii) an `S'-shaped trajectory with one turning point. For both trajectory types, the radius of rotation is uniformly sampled between 1 and 1.5 units, and the corresponding rotation angle is set between 90° and 180°. The object is animated to follow this trajectory over 200 steps while maintaining a fixed rotation center to simulate rotational motion. The movements are rendered at 5 fps with 32 keyframes using Blender’s Cycles engine, with each object sampled between $1 \sim 8$ times. \cref{data_show} presents visualizations of the animated data with various rotational trajectories and objects.

\subsection{Ablation Study on Data scale}
\begin{figure}[t]
\centering
{
\includegraphics[width=0.95\columnwidth]{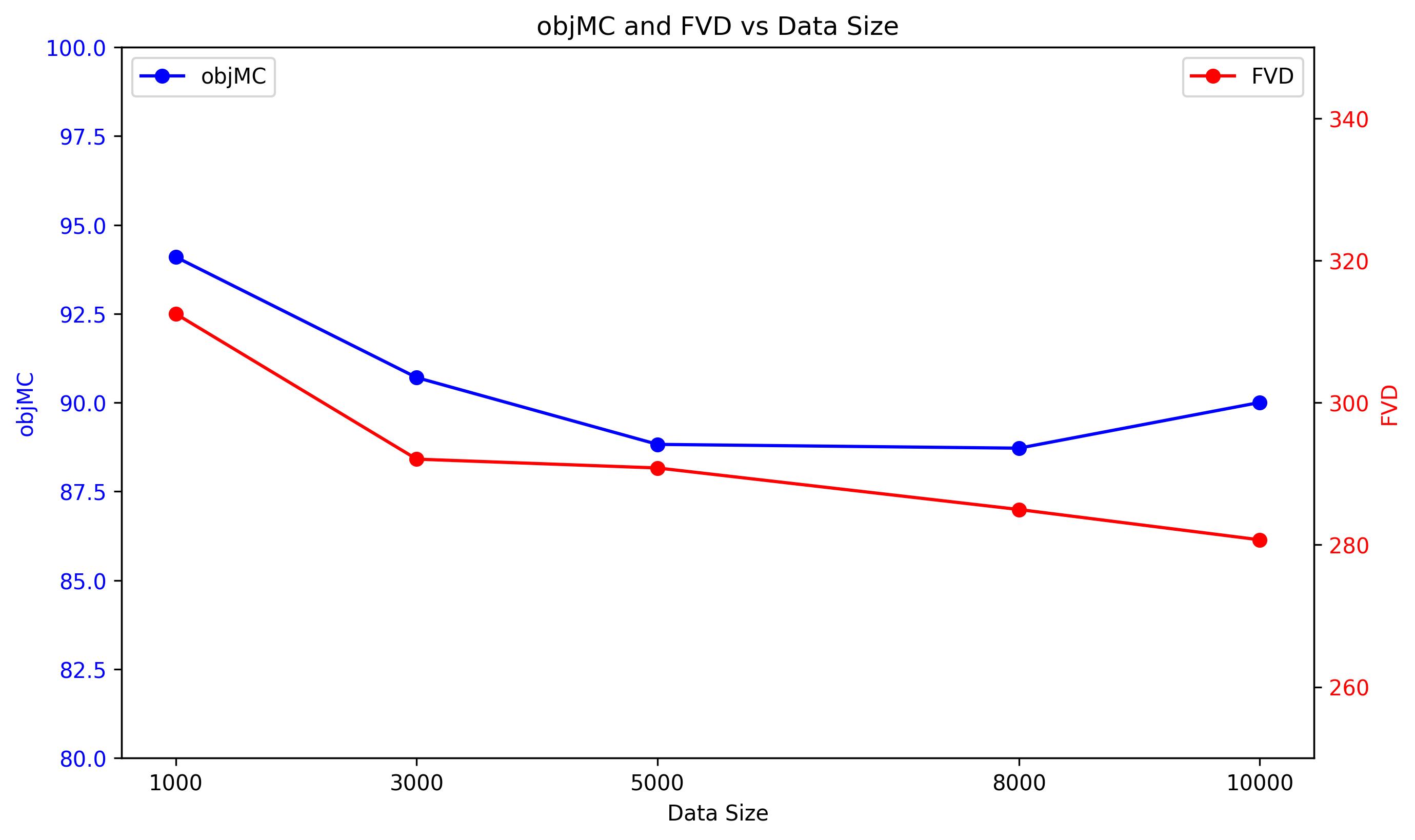} }
\vspace{-2mm}
\caption{
Visualization of objMC and FVD with scaling diversity.}
\label{ab_size_2}
\vspace{-0.2cm}
\end{figure}

To ensure sufficient diversity and robustness for pretraining, we carefully evaluated our dataset of 2,000 objects with sampled videos, determining that this amount approaches the model’s capacity limit, as shown in \cref{ab_size,ab_size_2}.

\begin{table}[!htb] \tiny
\centering
\caption{Comparison results on the synthetic validation set of our designed ablation studies on synthetic data scale during the pretraining stage.}
\vspace{-2mm}
\renewcommand{\arraystretch}{1.1}
\resizebox{0.9\linewidth}{!}{
\begin{tabular}{ccccc}
\hline
\multicolumn{1}{c}{\multirow{2}{*}{Vids}} & {\multirow{2}{*}{Objs}} & \multicolumn{3}{c}{Synthetic-Val}  \\ \cline{3-5}
\multicolumn{2}{c}{}                                  & ObjMC$\downarrow$           & FID $\downarrow$       & FVD$\downarrow$                    \\ \hline
1,000&200                                         & 0.1987         & 48.05         & 190.35                    \\ 
2,000 & 400                                      & 0.2065 &47.19         & 187.12                     \\
5,000 &1,000                                        & 0.1960& 46.62        & 185.47                          \\
10,000 &2,000                                           &    0.1895 &   46.34   & 186.01                          \\ \hline
\end{tabular}
}
\label{ab_size}
\end{table}

\noindent \textbf{Ablation Study on syntehtic data scale.}
To ensure effective pretraining as a pose-aware 3D understanding injection, it is critical to collect a sufficient amount of data for robust model learning. 
To investigate the optimal data size for two-stage pose-aware pretraining, we conducted an ablation study examining the impact of the number of training videos and the corresponding rendered objects.
Specifically, we extended the number of videos from 1,000 to 10,000, scaling the quantity of rendered objects accordingly. As presented in \cref{ab_size}, the model demonstrates a reliable rotational understanding with 5,000 or more training videos, whereas it struggles to learn diverse rotational patterns with only 1,000 or 2,000 videos. 
Furthermore, the performance difference between 5,000 and 10,000 videos is negligible, suggesting that 5,000 training videos are sufficient for the 3D-aware pretraining stage. Collecting data beyond 10,000 videos appears to offer no significant advantage and is both unnecessary and inefficient.

Additionally, as demonstrated in \cref{ab_size_2}, the trained model benefits from increased data diversity, where we shortened training steps due to time constraints.

\section{More Visualization Results}
\begin{figure*}[t]
    \centering
    \includegraphics[width=0.99\textwidth]{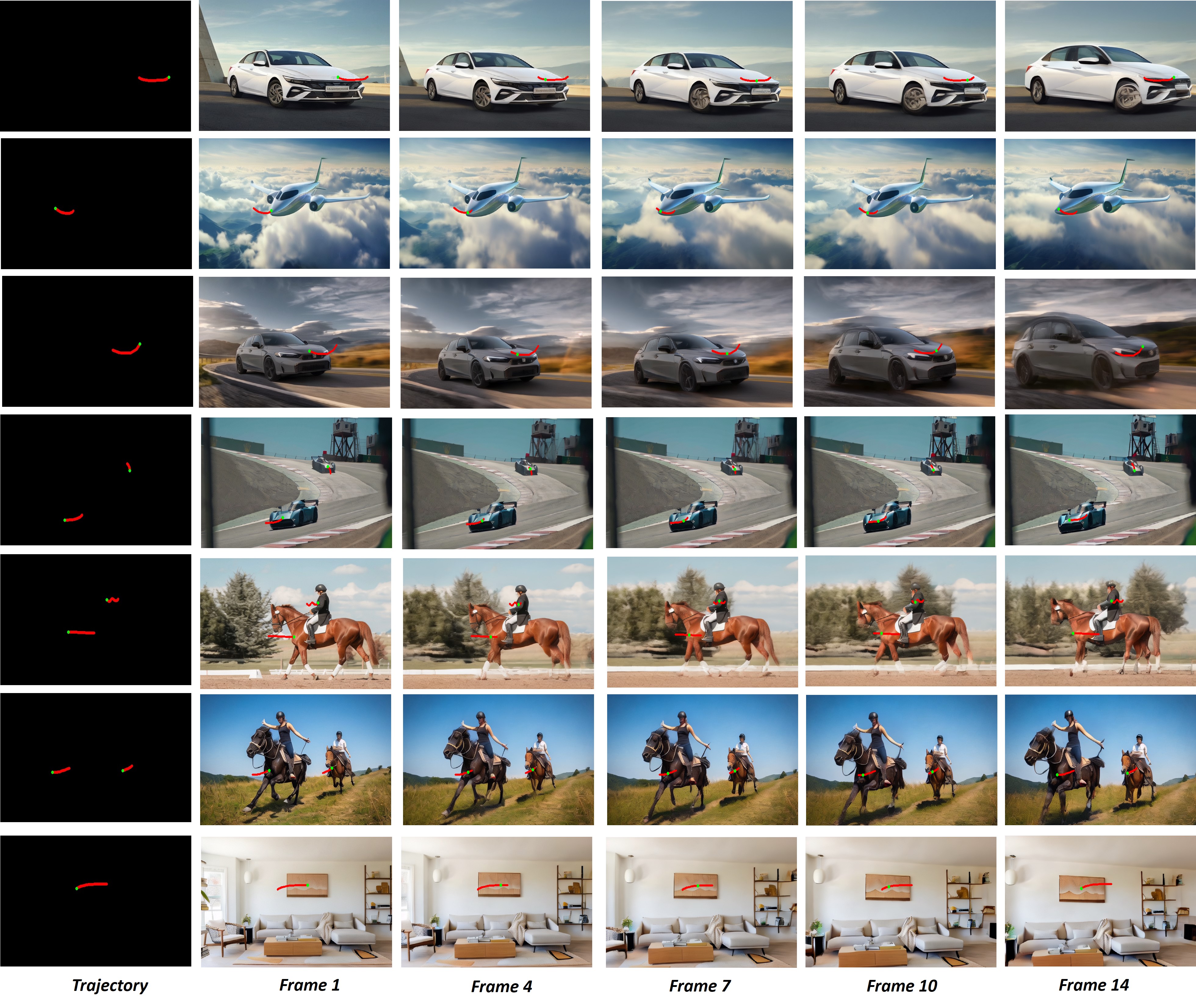} 
    \vspace{-3mm}
    \caption{More visualization results of our \shortname facing various rotational trajectories for single-object and multiple-object, and overall camera controlling.}
    \label{visual_res_spp}
\end{figure*}

In this section, we provide more visualization results from our model in \cref{visual_res_spp}, including different rotational trajectories for single-object and multiple-object controlling, and the overall camera controlling.

It can be observed that our model generates both precise rotational and translational motion following trajectories for various objects and also maintains superior object entity and video quality with potential wide-range motions.  

\section{Visualization Results for Ablation Studies}
In this section, we present additional visualization results (\cref{ab_vipsg_supp,ab_blender_supp}) to complement the metric-based experiments for our ablation studies discussed in the main paper (Sec. 5.2), including the ablation performance on open-domain videos and the pretraining stage using synthetic data.
\subsection{Performance on Open-Domain Dataset}
\begin{figure*}[t]
    \centering
    \includegraphics[width=0.99\textwidth]{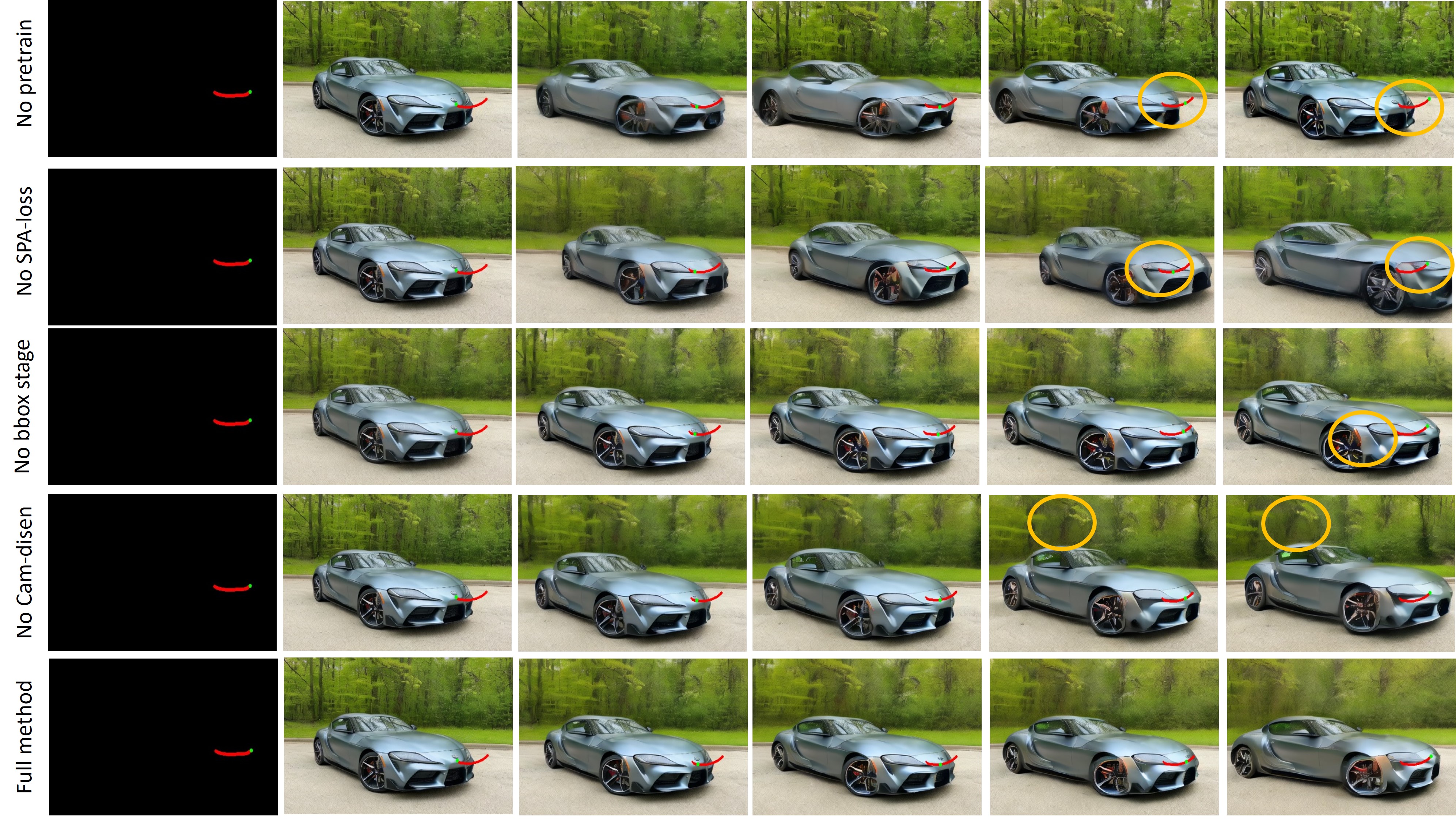}
    \caption{Visualization for generated results of ablation study on open-domain videos.}
    \label{ab_vipsg_supp}
\end{figure*}

\cref{ab_vipsg_supp} shows ablation results on the open-domain dataset. 
The model trained without two-stage pretraining exhibits poor trajectory-following capability in later frames, demonstrating a lack of temporal consistency. 
For the model trained without bounding box supervision (`No bbox stage'),
the pose changes in the generated rotational motions under wide-range motion scenarios are notably less pronounced compared to the final model.
Additionally, removing the spatial enhancement loss during training leads to a collapse in object identity, resulting in poor visual coherence. 
While the model trained without the camera disentanglement module retains comparable pose-aware generation capability for rotational and accurate motions, it suffers from misaligned camera perspectives and increased instability during inference, leading to frequent camera movements that degrade the overall quality.

\subsection{Performance on Synthetic Dataset}
\begin{figure*}[t]
    \centering
    \includegraphics[width=0.99\textwidth]{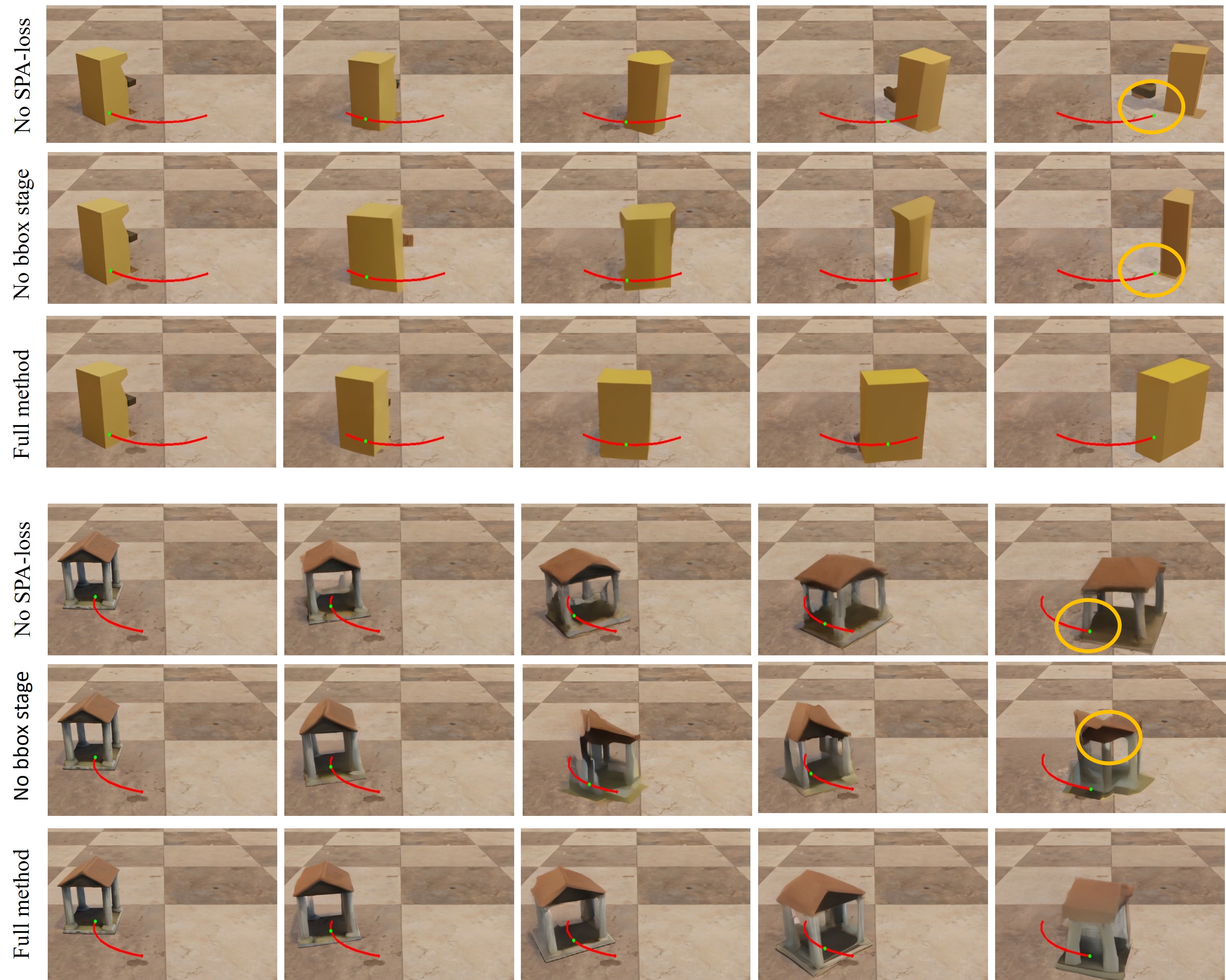}
    \caption{Visualization for the generated results of ablation study on the synthetic dataset.}
    \label{ab_blender_supp}
\end{figure*}

We further present additional visualizations to support the ablation study conducted during our designed pretraining stage. As demonstrated in \cref{ab_blender_supp}, the model trained without spatial enhancement loss exhibits a notable degradation in trajectory-following accuracy. Additionally, the model that omits first-stage pretraining with 3D bounding boxes experiences significant object collapse in the final few frames, accompanied by corresponding worse motion accuracy.

\section{Implementation Details}
\parag{Training details.}
Our full training pipeline includes three stages: two-stage pose-aware pre-training on a synthetic dataset and final-stage camera-disentangled finetuning on open-domain videos. All training is deployed on a single A100, requiring about 40G memory usage for the batch size of 1. Each stage of the pre-training took 5k steps using an AdamW optimizer with a 1e-5 learning rate. Our final finetuning took another 10k steps. 

\parag{Real-world video annotation.} 
To annotate the real-world video with trajectory and camera poses, we exploit two separate steps.
For the trajectory, following DragAnything~\cite{wu2024draganything}, we compute the center locations of objects based on their corresponding instance masks in our selected dataset. 
We then employ CoTracker2~\cite{karaev23cotracker} to extract the motion trajectories of these center points, which serve as conditional input. To ensure consistency between synthetic and real-world data, the extracted trajectories are formatted as discrete pixel space points as in our synthetic dataset.
As for the camera pose, we extract camera parameters from videos through DROID-SLAM \cite{teed2021droid}. 

\parag{Trajectory sampler.} 
For enhanced robustness during inference, we modify the original trajectory at the object center by sampling trajectories more sparsely within the projected 2D bounding boxes, with n sampled dragging initial points ($n \leq 8$). These sampled points are dragged along the original trajectory’s motion path, and their movements are visualized as images for further training.

\section{Discussion}
\subsection{Why Using 3D Poses as Supervision Signal}
In the context of 3D-aligned video generation, various signals, such as depth or depth heatmaps, encode potential 3D information. Compared to using depth as an internal supervision signal, 3D bounding boxes provide explicit object-level pose and approximate location, significantly improving object appearance refinement. Another possible approach is to introduce an additional depth dimension during generation. However, unlike 2D object localization, accurately estimating depth during inference remains highly challenging, often resulting in severe mismatches at the inference stage.

\subsection{Limitation and Future Work}
Through our experiments, we identified three primary limitations of the current model:
\begin{itemize}
    \item \textbf{Limited capability for wide-range rotations of dynamic objects}. 
    While the model can generate stable and accurate pose-aware rotational motions for \emph{static} objects such as cars, planes, and horses, it struggles with wide-range rotations for \emph{dynamic} objects, such as humans. This limitation primarily stems from the lack of rotationally dynamic objects, such as people or animals, in the training dataset. 
    A potential solution is to incorporate additional animatable objects, such as walking bears or running avatars, into the synthetic dataset during pretraining.

    \item \textbf{Insufficient camera control capacity}. 
    Despite employing a camera-disentanglement module to enhance object-centric trajectory understanding, the current module fails to provide precise camera control. This issue could be addressed by incorporating large-scale camera-specific datasets, such as RealEstate10K \cite{zhou2018stereo}, for further training or by directly integrating precisely controlled moving cameras into the synthetic dataset during the pretraining phase.

    \item \textbf{Blurry background in large motions}.
    Our model may generate a blurry background and this problem comes from the following two aspects. Firstly, the base SVD model struggles to maintain a consistent background when handling large motions. Secondly and most importantly, while our model improves trajectory-matching accuracy for large motions, the inherent blurriness in large movements from training data negatively impacts overall performance. This problem can be mitigated by fine-tuning on high-quality datasets.

\end{itemize}

\end{document}